\documentclass[11pt]{article}

\usepackage[preprint]{acl}

\usepackage{times}
\usepackage{latexsym}
\usepackage{todonotes}
\usepackage[T1]{fontenc}

\newcommand{\projname}{\textsc{AmbiSteer}}

\usepackage[utf8]{inputenc}

\usepackage{microtype}

\usepackage{inconsolata}

\usepackage{graphicx}
\usepackage{amsmath}
\usepackage{booktabs}
\usepackage{tabularx}
\usepackage{multirow}
\usepackage[table]{xcolor}

\usepackage[most]{tcolorbox}

\newtcolorbox{judgepromptbox}[1]{
    enhanced,
    breakable,
    colback=black!2,
    colframe=black!35,
    boxrule=0.4pt,
    arc=1mm,
    left=1.5mm,
    right=1.5mm,
    top=1.5mm,
    bottom=1.5mm,
    fonttitle=\bfseries,
    title={#1}
}

\newtcolorbox{promptbox}[1]{
    enhanced,
    colback=black!2,
    colframe=black!35,
    boxrule=0.4pt,
    arc=1mm,
    left=1.5mm,
    right=1.5mm,
    top=1.5mm,
    bottom=1.5mm,
    fonttitle=\bfseries,
    title={#1}
}

\newcommand{\ws}[2]{%
    \ifnum#1=0
        \multicolumn{1}{c}{#2}%
    \else
        \multicolumn{1}{c}{\cellcolor{cyan!#1}{#2}}%
    \fi
}

\title{\projname: Selectively Eliciting Ambiguity-Aware Responses \\ in Vision-Language Models}

\author{
  \textbf{Monica Riedler\thanks{Equal contribution.}\textsuperscript{1,2}} \quad
  \textbf{Samuele Antonelli\footnotemark[1]\textsuperscript{3}} \quad
  \textbf{Barbara Plank\textsuperscript{1,2}} \quad
  \textbf{Raffaella Bernardi\textsuperscript{4}}
  \\
  \textsuperscript{1}MaiNLP, Center for Information and Language Processing, LMU Munich, Germany \\
  \textsuperscript{2}Munich Center for Machine Learning (MCML), Munich, Germany \\
  \textsuperscript{3}University of Trento,
  \textsuperscript{4}Free University of Bozen-Bolzano, Italy \\
  \small{
  \texttt{\{monica.riedler,b.plank\}@lmu.de} \quad
  \texttt{samuele.antonelli@unitn.it} \quad
  \texttt{raffaella.bernardi@unibz.it}
  }
}

\begin{document}
\maketitle
\begin{abstract}

Vision-language models (VLMs) often exhibit overconfidence, providing firm answers even when visual evidence is insufficient. A prominent case is referential ambiguity in Visual Question Answering (VQA), where multiple entities in an image are plausible referents of a question. 
Addressing such cases requires recognizing when ambiguity is present: ambiguous questions can be handled by asking for clarification or enumerating plausible referents, while unambiguous questions should receive a direct answer. Current VLMs struggle with this distinction, often defaulting to a single interpretation.
To investigate whether activation steering can serve as an inference-time mechanism to elicit such behaviors, we introduce \projname{}, a lightweight framework to selectively elicit ambiguity-aware behavior.
Across three VLMs, we find that steering alone has limited and model-dependent effects, while prompting is generally more effective. Combining prompting and steering substantially increases clarification and enumeration behaviors, but also elicits them on unambiguous cases. \projname{} addresses this using a linear classifier to predict input ambiguity and condition the steering direction accordingly, increasing the proportion of inputs handled appropriately from 59.0\% to 77.0\% on average across models. Our work sets the ground for lightweight approaches to reduce model overconfidence in uncertain multimodal settings.

\end{abstract}

\begin{figure}[th!]
    \centering
    \includegraphics[width=0.95\linewidth]{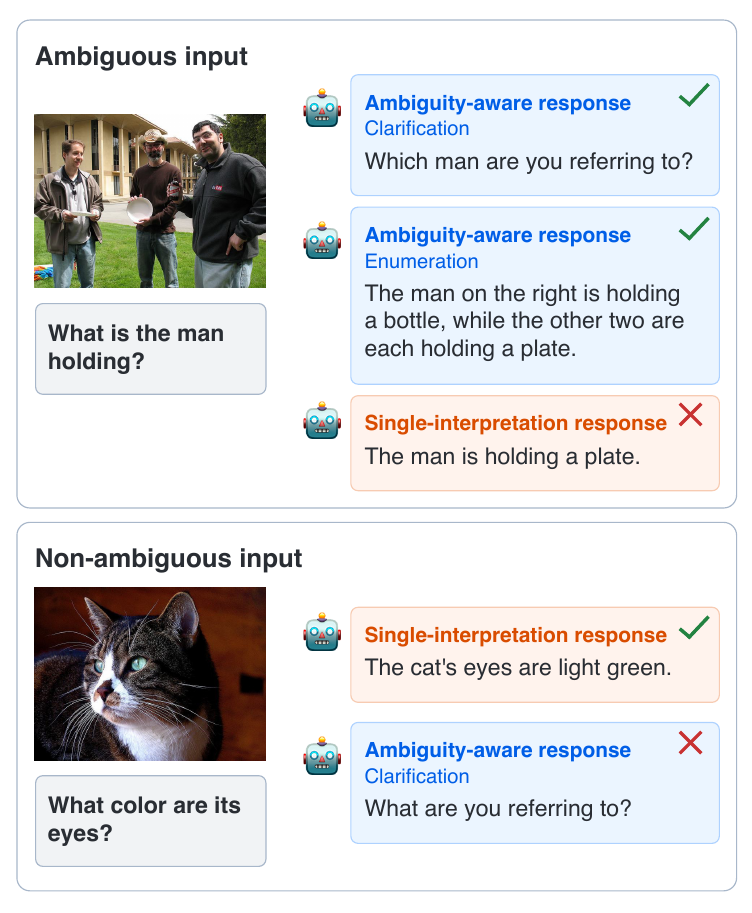}
    \caption{Referentially ambiguous questions call for ambiguity-aware responses, such as asking for clarification or enumerating multiple plausible referents, whereas non-ambiguous questions call for a direct, single-interpretation response.
}
    \label{fig:selective_ambiguity}
\end{figure}

\section{Introduction}
Visual Question Answering (VQA) requires models to interpret natural-language questions in visual context \citep{Antol_2015_ICCV}. This often involves resolving which entity in the image a question refers to. However, visual context does not always uniquely identify the intended referent. For example, in Figure~\ref{fig:selective_ambiguity}, the question ``What is the man holding?'' is referentially ambiguous because multiple people in the image are equally plausible referents. Motivated by \citet{testoni_racquet_2025}, we focus on two ambiguity-aware strategies observed in human responses to referential ambiguity: asking which referent is intended (\emph{clarification}) and providing answers for multiple plausible referents (\emph{enumeration}). Yet, their analysis shows that current VLMs often overcommit to a single arbitrary interpretation without acknowledging ambiguity.

Crucially, clarification and enumeration should not be elicited unconditionally. Figure~\ref{fig:selective_ambiguity} also illustrates the complementary case: when the referent is clear, a direct answer is most appropriate, while clarification or enumeration would be unnecessary and violate Grice's maxim of quantity by providing more information than required \citep{grice1975}. Effective ambiguity handling is therefore inherently \textit{selective}: ambiguity-aware responses are appropriate when the referent is unresolved, but not when the intended referent is already clear.

Two recent works address referential ambiguity through task-specific model training: ClearVQA trains models to ask clarification questions for ambiguous visual questions \citep{jian_teaching_2025}, while AQuA trains models to adapt their response strategy to different levels of ambiguity \citep{jang_aqua_2026}.
Explicit prompting provides a training-free alternative, but is not consistently effective across models \citep{testoni_racquet_2025, jang_aqua_2026}. 
We investigate activation steering as a lightweight inference-time control mechanism that leaves the VLM's weights unchanged. By modifying internal representations during generation, activation steering offers a way to selectively promote or suppress target behaviors without fine-tuning.

To this end, we introduce \projname, a framework for selective ambiguity handling in VQA. Using ClearVQA and AQuA to study clarification and enumeration behaviors, respectively, we derive steering vectors from contrastive pairs consisting of an ambiguity-aware response and a response that commits to a single interpretation. 
Because effective ambiguity handling requires knowing both \textit{how} to respond and \textit{when} to intervene, \projname{} combines activation steering with  classifier-gated ambiguity prediction. A linear classifier over VLM representations predicts whether each image-question pair is ambiguous, and its prediction determines the steering direction, encouraging ambiguity-aware behavior for ambiguous inputs and suppressing it otherwise.

Experiments across three VLMs show that steering alone has limited effects, while prompting is generally stronger but varies considerably across models. Interestingly, combining prompting and steering consistently increases ambiguity-aware behavior beyond either method alone, in many cases by a large margin. This tendency, though, extends to non-ambiguous inputs, where clarification or enumeration is undesirable.
\projname{} improves this balance and achieves the best trade-off between eliciting ambiguity-aware responses when needed and avoiding them when the referent is clear, increasing the proportion of appropriately handled inputs from 59.0\% for combined prompting and steering to 77.0\% on average.

\section{Related Work}
\label{sec:related}
\paragraph{Referential Ambiguity in VQA.}
Ambiguity as a resolvable property of a question has been studied in text-only open-domain QA, where \citet{min_ambigqa_2020} introduced AmbigQA to resolve multiple valid answers through rewrites. Early multimodal work followed a similar direction: \citet{stengel_eskin_chicken_2023} identified ambiguous VQA questions via annotator disagreement and trained models to rephrase them, while \citet{prasad_rephrase_2023} used Large Language Models (LLMs) to rephrase underspecified questions in a zero-shot setting.
Recent benchmarks explicitly target referential ambiguity in visual scenes, highlighting that standard VLMs struggle when visual context leaves the referent underspecified \citep{testoni_racquet_2025, Nam_2025_ICCV}.
More recent studies train models to handle ambiguity through clarification questions or referent enumeration \citep{jian_teaching_2025, jang_aqua_2026}. In contrast, we selectively elicit both behaviors at inference time without modifying model weights.

\begin{figure*}[th!]
    \centering
    \includegraphics[width=\textwidth]{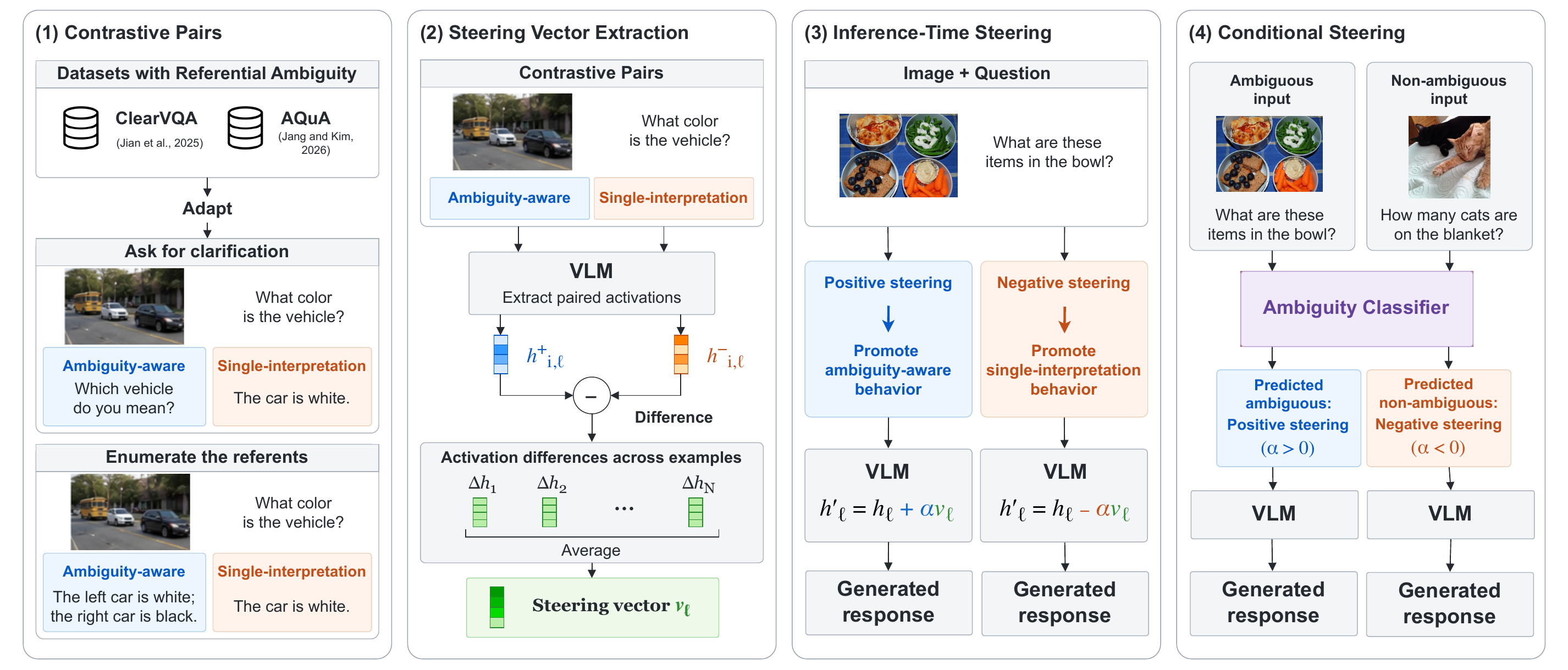}
    \caption{Overview of the \projname{} framework. We adapt two datasets to obtain ambiguity-aware and single-interpretation contrastive pairs, extract steering vectors from their activation differences, and apply steering at inference time. Conditional steering uses an ambiguity classifier to select the steering direction.}
    \label{fig:pipeline}
\end{figure*}

\paragraph{Linear Separability and Activation Steering.}
Linear probing studies suggest that high-level concepts are often organized along linearly separable directions in the activation space, supporting the linear representation hypothesis \citep{alain_understanding_2016, marks2024the, park_linear_2023}. Building on this, activation steering modifies a model's internal representations at inference time to elicit or suppress specific behaviors without updating model parameters. Early approaches, such as Activation Addition and Inference-Time Intervention \citep{turner_steering_2023, li2024inferencetimeinterventionelicitingtruthful, zou_representation_2023}, derive steering directions from activation differences obtained from a single contrastive pair or from free-form text exhibiting the target behavior. Contrastive Activation Addition (CAA; \citealp{rimsky_steering_2024}) instead constructs steering vectors from a dataset of minimal pairs, averaging the resulting activation differences across all pairs to obtain a less noisy direction.
A key limitation of unconditional steering is that it intervenes regardless of whether steering is needed. Most closely related to our work, conditional Activation Steering (CAST; \citealp{lee2025programmingrefusalconditionalactivation}) addresses this by learning a condition direction and steering only when a criterion is met.
Recent VLM steering studies focus mainly on object hallucination mitigation \citep{yin_dynamic_2026, su_activation_2025}, instruction following \citep{sivakumar_steervlm_2025}, and configurable refusal behavior \citep{yang_steering_2026}, extending text-only findings \citep{arditi_refusal_2024} to the visual domain. However, existing work has not addressed visual ambiguity, nor principled criteria for deciding when steering for ambiguity should be applied. Our approach follows the premise of conditional steering, but addresses this gap by using an internal ambiguity classifier to dynamically determine whether to promote or suppress ambiguity-aware behavior.

\section{\projname{} Framework}

Figure~\ref{fig:pipeline} summarizes our proposed \projname{} framework, described in the following subsections.

\subsection{Contrastive Pairs}
\label{sec:task-datasets}

To study the two central disambiguation behaviors, we build contrastive pairs using two distinct datasets: ClearVQA \citep{jian_teaching_2025}, which investigates clarification-seeking behavior, and AQuA \citep{jang_aqua_2026}, which studies enumeration behavior. Because these benchmarks are not designed for representation-based intervention, for each dataset we generate paired responses for identical image-question contexts, contrasting an ambiguity-aware response against a single-interpretation response that arbitrarily commits to a specific referent. These contrastive pairs serve as the training signal for deriving steering vectors as well as evaluating selective ambiguity handling. Complete descriptions of the dataset adaptation and examples of resulting contrastive pairs are detailed in Appendices ~\ref{app:dataset_adaptation} and ~\ref{app:constrastive_constructions}, respectively. 

\subsection{Steering Vector Extraction}

\paragraph{Contrastive constructions.}

We seek a direction in activation space that captures the distinction between ambiguity-aware and single-interpretation behavior, and consider two ways of constructing this contrast.
The first follows \citet{arditi_refusal_2024}, where we form two sequences by appending the ambiguity-aware and single-interpretation responses separately to the same image–question context and extract the resulting activations. We refer to this as the free-form (FF) construction because the contrast is formed directly from the open-ended responses. This closely matches the setting in which steering is ultimately applied. However, the two responses may differ substantially in length, wording, and lexical content, so their activation difference may capture incidental linguistic variation in addition to the intended response strategy.

To reduce this source of variation, we additionally consider Contrastive Activation Addition (CAA; \citealp{rimsky_steering_2024}), where the two responses are presented as options in a shared multiple-choice formatted prompt and the contrasted sequences differ only in the final answer-letter token . CAA therefore provides a more controlled contrast, but introduces a format gap between extraction and generation, since the direction is derived from multiple-choice responses and later applied to open-ended generation. We therefore test both constructions, since it is unclear whether a cleaner contrast or a closer match to open-ended generation is more important for effective steering.

\paragraph{Steering vector computation.}
For either construction, let \(h^{+}_{i,\ell}\) and \(h^{-}_{i,\ell}\) denote the extracted activations for example \(i\) at decoder layer \(\ell\). We compute the steering vector as the mean activation difference across \(N\) examples:
\[
v_\ell =
\frac{1}{N}
\sum_{i=1}^{N}
\left(
h^{+}_{i,\ell} - h^{-}_{i,\ell}
\right)
\]
We derive vectors independently for each decoder layer, model, and dataset.

\subsection{Inference-Time Steering}

At inference time, we add the steering vector to the decoder residual stream at each generation step. At decoder layer $\ell$, we modify the hidden state as  
$
h'_\ell = h_\ell + \alpha v_\ell,
$ 
where $v_\ell$ is the steering vector extracted at that layer and $\alpha$ controls the strength and sign of the intervention. Positive coefficients are designed to promote the ambiguity-aware behavior, while negative coefficients are designed to suppress it; $\alpha = 0$ corresponds to generation without steering.

We apply steering at either a single layer or across a contiguous set of layers (more details are provided in Section \ref{sec:hyperparameters} and Appendix \ref{app:layer_coeff_selection}). In the multi-layer setting, the corresponding layer-specific vector $v_\ell$ is added at each selected layer.

\subsection{Conditional Steering}
\label{sec:conditional_steering}

The intervention described above specifies how to promote or suppress ambiguity-aware behavior, but not when each is appropriate: applying positive steering indiscriminately can trigger unnecessary clarification or hallucinated referents on non-ambiguous inputs. We therefore condition the steering intervention on predicted input ambiguity, using a linear probe trained to classify each image-question pair as ambiguous or non-ambiguous.

\section{Experimental Setup}

\subsection{Data Splits and Models}
\label{sec:data_splits}

\begin{table}[h]
\centering
\footnotesize
\setlength{\tabcolsep}{3.5pt}
\renewcommand{\arraystretch}{1.08}

\begin{tabular}{@{}llrrr@{}}
\toprule
\textbf{Dataset} & \textbf{Split} & \textbf{$N$} & \textbf{Ambig.} & \textbf{Non-ambig.} \\
\midrule

\multirow{3}{*}{ClearVQA}
& Train & 728 & 395 & 333 \\
& Dev   & 300 & 300 & -- \\
& Test  & 600 & 300 & 300 \\

\midrule

\multirow{4}{*}{AQuA}
& Train (steering)   & 900 & 900 & -- \\
& Train (classifier) & 900 & 450 & 450 \\
& Dev                & 300 & 300 & -- \\
& Test               & 600 & 300 & 300 \\

\bottomrule
\end{tabular}

\caption{Data splits used in our experiments.}
\label{tab:data-splits}
\end{table}

\paragraph{Data Splits.} Table~\ref{tab:data-splits} summarizes our data splits.
Since training data is more constrained in ClearVQA, we use a single split containing ambiguous and non-ambiguous samples for both extracting steering vectors and training the ambiguity classifier. In contrast, the larger scale of AQuA allows us to use an exclusively ambiguous split to obtain the steering vectors. Since ambiguity classification requires a balanced supervisory signal across both conditions, we allocate a distinct split with an equal number of ambiguous and non-ambiguous examples to train the classifier. Full details on the dataset structure and adaptation are provided in Appendix~\ref{app:dataset_adaptation}.
\footnote{Code and data will be made available upon acceptance.} 

\paragraph{Models.} 
We evaluate three open-weight VLMs of comparable scale: Qwen3-VL-8B-Instruct~\citep{bai_qwen3-vl_2025}, InternVL3.5-8B~\citep{wang_internvl35_2025}, and Molmo2-8B~\citep{clark_molmo2_2026}. 
Steering vectors are extracted separately for each model.
Generation settings are kept fixed across experimental conditions for a given model. Additional model and decoding details are provided in Appendix~\ref{app:implementation_details}.

\subsection{Generation Settings}

We compare \projname{} with four generation settings that vary whether activation steering is applied and whether a task-specific prompt instruction is provided. In addition, we report a Random baseline as lower bound and an Oracle as upper bound using gold ambiguity labels.

\paragraph{Baseline.}
We generate responses using the original model without an additional task-specific prompt or activation steering. This measures the model's default response behavior.

\paragraph{Steer only.}
We apply activation steering without providing an additional prompt instruction. This isolates the effect of the steering intervention itself.

\paragraph{Prompt only.}
We provide an explicit task instruction but do not modify the model's activations. For ClearVQA, the prompt asks the model to seek clarification if the intended referent is unclear. For AQuA, it asks the model to enumerate plausible referents when more than one interpretation is possible. Exact prompts are provided in Appendix~\ref{app:generation_instructions}.

\paragraph{Prompt + steer.}
We combine the task-specific prompt with activation steering. This tests whether steering provides additional gains and control over response behavior beyond prompting alone.

\subsection{Ambiguity Classification}
\label{sec:splits_ambiguity_labels}

We derive binary ambiguity labels separately for each dataset, leveraging human annotations. 
Details and examples are provided in Appendix \ref{app:human_ambiguity_annotation}.

Using the obtained labels, for each dataset we train a separate logistic-regression classifier to predict whether an image-question pair is ambiguous. The classifiers take as input the Qwen3-VL-8B-Instruct residual-stream activation at the final prompt token. We train one classifier per decoder layer and select the layer with the highest 5-fold cross-validation accuracy on the training data, yielding layer 22 for ClearVQA and layer 23 for AQuA. The selected classifiers achieve 77.2\% and 93.8\% test accuracy for ClearVQA and AQuA, respectively, and their predictions determine the steering sign at inference time across all three VLMs used for steering. Additional classifier details and comparisons are provided in Appendix~\ref{app:alt_classifiers}.

\subsection{Evaluation Metrics}
\label{sec:metrics}

We evaluate two central aspects: whether the interventions elicit the intended ambiguity-aware behavior, measured by clarification and enumeration rate, and whether this behavior is applied selectively, measured by behavioral accuracy. We additionally report a token-level degeneration rate to monitor repetitive generation under steering.

\paragraph{Clarification and enumeration rate.}
We measure how often generated responses exhibit the target behavior, i.e., asking a relevant clarification question on ClearVQA or considering multiple plausible referents and answering for each on AQuA. We use Meta-Llama-3.1-8B-Instruct~\citep{grattafiori2024llama3herdmodels} as a few-shot judge based on the original question and generated response.
Exact evaluation prompts and few-shot demonstrations are provided in Appendix~\ref{app:llm_judge_prompts}.

\paragraph{Behavioral accuracy.}
Clarification and enumeration rates measure how frequently the target behavior occurs, but not whether it occurs in the appropriate cases. We therefore report behavioral accuracy, counting a response as behaviorally correct when the target behavior is present on an ambiguous input and absent on a non-ambiguous input.

\paragraph{Degeneration rate.}
We measure whether steering causes excessive repetition. A response is considered degenerate if any 4-gram occurs at least eight times. Degenerate generations are counted as behaviorally incorrect.

\subsection{Human Validation}
\label{sec:human_validation}

\paragraph{LLM-judge validation.}
To validate the LLM-based clarification and enumeration rates, two annotators independently label 100 generations from each dataset using the same criteria as the LLM judge.
On ClearVQA, the annotators agree on 97\% of examples ($\kappa = 0.93$), while the LLM judge agrees with the individual annotators on 90-91\% of examples. On AQuA, human agreement is 91\% ($\kappa = 0.80$), and LLM-human agreement is 89-90\%. Full details of the annotation procedure, annotators, and agreement scores are shown in Appendix \ref{app:llm_judge_validation}.

\paragraph{Factual consistency.}
For AQuA, the enumeration metric captures whether a response considers multiple plausible referents, but not whether the information provided for those referents is supported by the image. We therefore conduct a separate human evaluation following the factual consistency criterion proposed by the authors of the AQuA dataset~\citep{jang_aqua_2026}. A human annotator is shown the image, question, and generated response and judge whether the response is grounded in the visual evidence. We report further details of the annotation procedure in Appendix \ref{app:factual_consistenc_annotation}.

\subsection{Steering Hyperparameter Selection}
\label{sec:hyperparameters}

We select steering layers and coefficients on the development sets using clarification or enumeration rates from open-ended generations. To limit per-condition tuning, we select a set of intervention layers using the multiple-choice CAA vectors, and reuse these layers for the free-form construction. We consider $k \in \{1,5\}$, where $k$ denotes the number of consecutive layers being steered. We sweep each candidate layer with coefficients $\alpha=+1$ and $-1$, compute the difference in target-behavior rate, and select the layer, or contiguous five-layer window, with the largest difference averaged across models and datasets. This selects layer~21 for $k=1$ and layers~20--24 for $k=5$. After fixing the layers, we tune the steering coefficient $\alpha$ separately for each model, dataset, vector construction, $k$, and steered generation setting. Positive and negative coefficients are selected to maximize and minimize the target-behavior rate, respectively.
Full sweep ranges, normalization details, and selected coefficients are provided in Appendix~\ref{app:layer_coeff_selection}.

\section{Results and Analysis}

We evaluate both free-form (FF) and multiple-choice (CAA) steering. Since CAA is more consistently effective across models and datasets, we focus on CAA with ($k=5$) in the main text for clarity. Appendix \ref{app:free_form} reports the corresponding FF results and ($k=1$) results for both constructions.

\subsection{Probing Ambiguity-Aware Behavior}

Prior work has used linear probes to examine whether attributes targeted by activation interventions are linearly represented in model activations \citep{li2024inferencetimeinterventionelicitingtruthful, shim-etal-2026-linear}.
We similarly assess whether the CAA vectors distinguish ambiguity-aware from single-interpretation responses. 
At each layer, the vector is defined by the difference between the mean training activations of the two response types. We use this direction as a linear probe, with the midpoint between the two means defining the decision boundary. As a control, we also measure separability by A/B answer label, allowing us to distinguish behavioral encoding from merely encoding the option letter.

\begin{figure}[t]
    \centering
    \includegraphics[width=\linewidth]{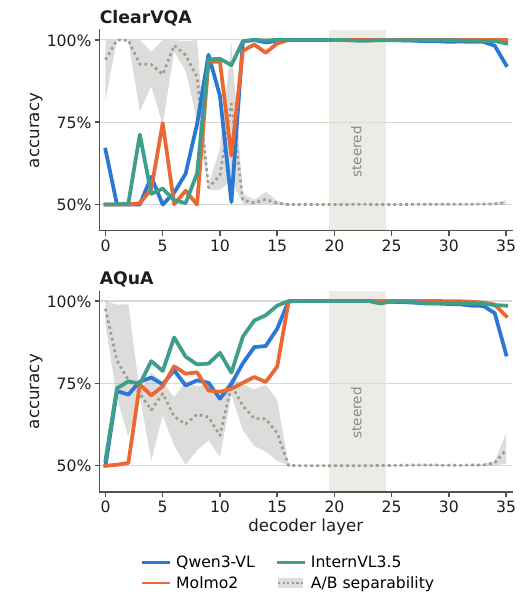}
    \caption{Layer-wise probing accuracy of CAA directions for response behavior and A/B label assignment on ClearVQA and AQuA. The A/B line shows the mean across models, with shading indicating the range. Vertical shading marks the steered layers (20–24).
    }
    \label{fig:linear_separability}
\end{figure}

Figure \ref{fig:linear_separability} shows a clear transition across layers. Across all three models and both datasets, response behavior becomes nearly perfectly separable between layers 12 and 16 and remains at 99.2–100\% accuracy across the steered layers. In contrast, A/B separability is high in early layers but drops to chance as behavioral separability emerges, reaching at most 50.2\% in the steered layers. This mirrors the transition from answer-letter to behavioral clustering observed by \citet{rimsky_steering_2024}, indicating that the CAA vectors used for intervention capture the intended response distinction rather than the A/B answer position.

\begin{figure*}[th!]
    \centering
    \includegraphics[width=0.96\textwidth]{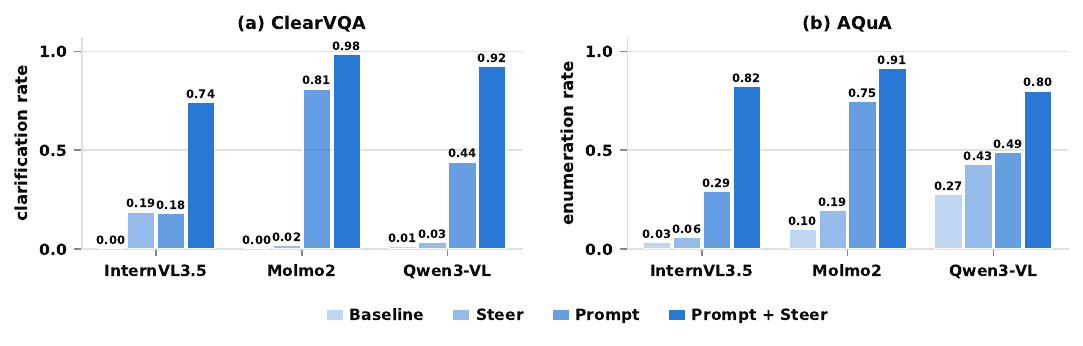}
    \caption{
        Target-behavior rates on ambiguous test instances.
        Panel (a) reports clarification rates on ClearVQA and panel (b) enumeration rates on AQuA, with 300 ambiguous test examples per dataset.
    }
    \label{fig:elicit_behavior}
\end{figure*}

\begin{figure*}[t]
    \centering
    \includegraphics[width=\textwidth]{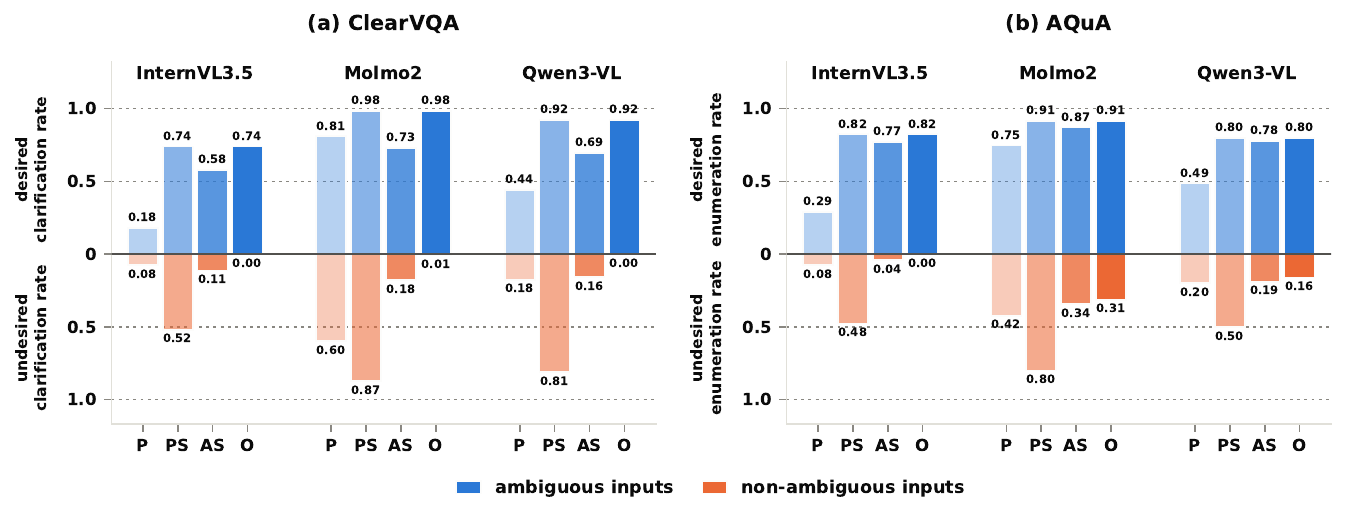}
    \caption{
    Clarification and enumeration rates on balanced test sets with 300 ambiguous and 300 non-ambiguous inputs per dataset.
    P = Prompt, PS = Prompt+Steer, AS = \projname{}, and O = Oracle.
    }
    \label{fig:selectivity}
\end{figure*}

\subsection{Eliciting Ambiguity-Aware Behavior}
We first evaluate whether activation steering and prompting elicit clarification or enumeration on ambiguous inputs.
Figure~\ref{fig:elicit_behavior} reports results on the 300 ambiguous test instances from each dataset.

Interestingly, \emph{steering alone provides only limited and uneven control} over the response behavior. On ClearVQA, clarification is essentially absent at baseline, and steer-only rates remain below 0.20 for all three models. On AQuA, the models enumerate more often, and steering produces somewhat larger shifts, particularly for Qwen3-VL. Overall, however, steering does not reliably elicit ambiguity-aware behavior across models and datasets.

Prompting is substantially more effective, although the models differ in how strongly they follow the instruction. Most importantly, \emph{combining prompting and steering is key:} it gives the highest clarification or enumeration rate for every model on both datasets. For Qwen3-VL on ClearVQA, for example, steering alone yields a clarification rate of 0.03 and prompting alone 0.44, while their combination reaches 0.92. The size of this additional gain varies by model: Molmo2 already responds strongly to prompting and gains less, while InternVL3.5 benefits much more from the combined intervention. 
For our ambiguity-handling tasks, steering therefore appears most effective as a complement to prompting,
reinforcing ambiguity-aware behavior elicited by the task instruction, while steering alone remains less reliable.
Our experiments with the free-form (FF) construction corroborate these findings (see Appendix \ref{app:free_form}).

\begin{table*}[th]
\centering
\small
\setlength{\tabcolsep}{6pt}
\renewcommand{\arraystretch}{1.08}

\newcommand{\accdeg}[2]{#1\,{\scriptsize (#2)}}

\begin{tabular}{llccccc}
\toprule
\textbf{Dataset} & \textbf{Model}
& \textbf{Prompt}
& \textbf{Prompt+Steer}
& \textbf{Random Gate}
& \textbf{AmbiSteer}
& \textbf{Oracle Gate} \\
\midrule
\multirow{3}{*}{ClearVQA}
& InternVL3.5
  & \accdeg{55.2}{0.0}
  & \accdeg{60.7}{0.5}
  & \accdeg{55.3}{0.3}
  & \accdeg{\textbf{73.2}}{0.0}
  & \accdeg{87.0}{0.2} \\
& Molmo2
  & \accdeg{60.7}{0.0}
  & \accdeg{55.5}{0.0}
  & \accdeg{52.4}{2.4}
  & \accdeg{\textbf{75.2}}{2.7}
  & \accdeg{96.2}{2.5} \\
& Qwen3-VL
  & \accdeg{63.0}{0.2}
  & \accdeg{51.3}{5.8}
  & \accdeg{49.0}{6.8}
  & \accdeg{\textbf{73.8}}{3.8}
  & \accdeg{92.7}{5.0} \\
\midrule
\multirow{3}{*}{AQuA}
& InternVL3.5
  & \accdeg{60.7}{0.0}
  & \accdeg{67.3}{0.3}
  & \accdeg{61.1}{0.2}
  & \accdeg{\textbf{86.5}}{0.3}
  & \accdeg{91.0}{0.3} \\
& Molmo2
  & \accdeg{66.0}{0.2}
  & \accdeg{55.5}{0.3}
  & \accdeg{56.1}{1.9}
  & \accdeg{\textbf{74.7}}{1.8}
  & \accdeg{77.8}{2.3} \\
& Qwen3-VL
  & \accdeg{64.2}{0.3}
  & \accdeg{63.5}{2.7}
  & \accdeg{63.2}{2.3}
  & \accdeg{\textbf{78.5}}{2.3}
  & \accdeg{81.2}{2.2} \\
\bottomrule
\end{tabular}

\caption{
Behavioral accuracy (\%) on the balanced test sets, with degeneration
rate (\%) shown in parentheses. Random Gate selects the steering
direction with probability 0.5 and is averaged over 200 random draws;
Oracle Gate uses the gold ambiguity label. Bold indicates the best
non-oracle behavioral accuracy.
}
\label{tab:behavioral-accuracy}
\end{table*}

\subsection{Selective Ambiguity Handling}

The previous analysis considers only ambiguous inputs, where increasing clarification or enumeration is desirable. A useful intervention, however, should promote these behaviors without inducing them on non-ambiguous ones. We therefore compare Prompt, Prompt+Steer, and \projname{} on balanced test sets of ambiguous and non-ambiguous inputs.
We omit Baseline and Steer, as they do not reliably elicit ambiguity-aware behavior.
Figure~\ref{fig:selectivity} shows the resulting scores separately for ambiguous and non-ambiguous inputs. Prompt+Steer strongly increases clarification and enumeration on ambiguous examples (86.4\% on average), but also substantially increases them on non-ambiguous inputs (66.3\%), where they are inappropriate. \projname{} reduces this undesired behavior to 17.0\% across all models and datasets, while maintaining a 73.6\% target-behavior rate on ambiguous inputs. The Oracle selects the steering direction from the gold ambiguity label, providing an upper bound.

To evaluate the overall appropriateness of model responses, Table~\ref{tab:behavioral-accuracy} reports behavioral accuracy, which rewards clarification or enumeration on ambiguous inputs and direct responses on non-ambiguous inputs.  \projname{} achieves the highest non-oracle behavioral accuracy across models and datasets, outperforming Prompt by 8.7 to 25.8 percentage points and Prompt+Steer by 12.5 to 22.5 points, as well as consistently outperforming the Random Gate. Degeneration remains low, with a maximum of 3.8\%.
 \projname{} comes close to Oracle performance on AQuA, while a larger gap remains on ClearVQA, consistent with the lower classifier accuracy on that dataset.

\subsection{Error Analysis}
\label{sec:error_analysis}

\paragraph{Sources of behavioral errors.}
To better understand the remaining failures under \projname, we separate degenerate generations from non-degenerate behavioral errors. We further distinguish errors associated with an incorrect ambiguity-classifier prediction from those that occur despite a correct prediction, which we refer to as classification and generation errors, respectively. As shown in Table~\ref{tab:error-analysis}, degeneration is rare, with at most 23 of 600 generations affected in any setting.
Behavioral errors are more frequent, and their source differs by dataset. On ClearVQA, most non-degenerate errors coincide with an incorrect ambiguity-classifier prediction, accounting for 72.0-95.5\% of errors across models, identifying ambiguity classification as the main bottleneck. On AQuA, the opposite pattern emerges: 58.2-83.0\% of errors occur despite a correct classifier prediction, indicating that the model often fails to produce the desired behavior even when the intended steering condition is applied.

\begin{table}[t]
\centering
\small
\setlength{\tabcolsep}{2.2pt}
\renewcommand{\arraystretch}{1.08}

\begin{tabular}{@{}llcccc@{}}
\toprule
\multirow{2}{*}{\textbf{Dataset}}
& \multirow{2}{*}{\textbf{Model}}
& \multirow{2}{*}{\textbf{Deg.}}
& \multicolumn{3}{c}{\textbf{Behavioral errors}} \\
\cmidrule(lr){4-6}
&
&
& \textbf{Total}
& \textbf{Class.}
& \textbf{Gen.} \\
\midrule

\multirow{3}{*}{ClearVQA}
& InternVL3.5 & 0  & 161 & 116 (72.0) & 45 (28.0) \\
& Molmo2      & 16 & 133 & 127 (95.5) & 6 (4.5) \\
& Qwen3-VL    & 23 & 134 & 127 (94.8) & 7 (5.2) \\

\midrule

\multirow{3}{*}{AQuA}
& InternVL3.5 & 2  & 79  & 33 (41.8) & 46 (58.2) \\
& Molmo2      & 11 & 141 & 24 (17.0) & 117 (83.0) \\
& Qwen3-VL    & 14 & 115 & 28 (24.3) & 87 (75.7) \\

\bottomrule
\end{tabular}

\caption{
Error breakdown for \projname{}. Degenerate generations (Deg.) are
reported separately and excluded from behavioral errors. Class. denotes
errors with an incorrect ambiguity-classifier prediction, while Gen.
denotes generation errors despite a correct prediction. Percentages are computed
over behavioral errors.
}
\label{tab:error-analysis}
\end{table}

\begin{figure}[th!]
    \centering
    \includegraphics[width=\linewidth]{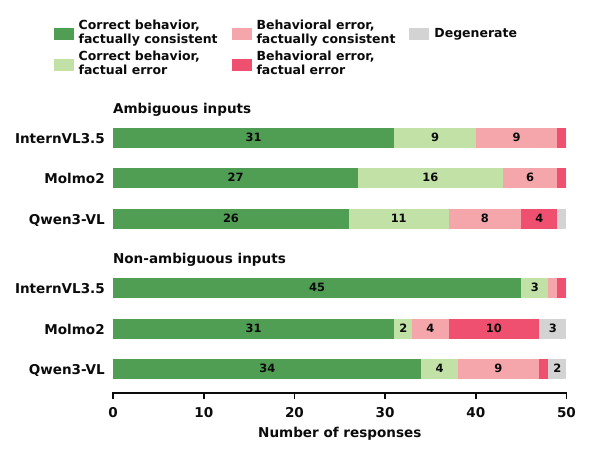}
    \caption{Behavioral and factual outcomes on AQuA.}
    \label{fig:error_cooccurrence}
\end{figure}

\paragraph{Behavioral Accuracy and Factual Consistency.}
Behavioral correctness measured by enumeration rate alone does not necessarily imply proper visual grounding. We therefore jointly analyze behavioral accuracy and human-rated factual consistency for 100 of the final AQuA outputs (Figure~\ref{fig:error_cooccurrence}). On ambiguous inputs, roughly half of behaviorally correct responses (26-31/50 across models) are also factually consistent, while a further 9-16/50  exhibit the correct enumeration behavior but contain factual errors. Thus, correct enumeration does not guarantee accurate description. Factual errors are less frequent on non-ambiguous inputs, though the overall error profile varies across models.
Representative generations illustrating both successful and failure cases are provided in Appendix \ref{app:qualitative_examples}.

\section{Conclusion}
We study whether activation steering can support selective handling of referential ambiguity in VLMs. Across three VLMs and two ambiguity-handling tasks, steering alone yields limited, model-dependent control, while prompting is generally more effective. Combining prompting and steering consistently yields the highest ambiguity-aware behavior rates, but also induces these behaviors on non-ambiguous inputs, where they are inappropriate.
By conditioning the steering direction on predicted ambiguity,  \projname{} improves this trade-off across all models and tasks, consistently increasing behavioral accuracy.
Our error analysis further shows that appropriate ambiguity handling does not ensure visual grounding, highlighting factual consistency as a key direction for future work.

More broadly, our findings place referential ambiguity within a wider challenge in current VLMs: a systemic overconfidence that leads models to provide firm answers even with insufficient evidence. By demonstrating that a classifier over model activations can dynamically gate steering vectors to impose pragmatic caution, 
\projname{} provides an inference-time solution for referential ambiguity. This lightweight paradigm could extend to broader settings where models must recognize uncertainty.

\section*{Limitations}
Our study has a number of limitations that also point to interesting directions for future investigation. First, the pool of evaluated models is restricted to three open-weight VLMs at a comparable scale (approximately 8B parameters), due to limited computational resources. It remains an open question whether our results, in particular the benefits of \projname{}, extend to different or larger models. Second, while our classifier achieves reasonable accuracy (77.2\% on ClearVQA and 93.8\% on AQuA), performance on ClearVQA leaves room for improvement, and our error analysis in Section \ref{sec:error_analysis} suggests that classifier errors can meaningfully affect the effectiveness of conditional steering. Third, we rely on a single model (Qwen3-8B) to generate the single-interpretation rephrasings used in our contrastive pairs. Although this provides a consistent procedure for constructing the pairs, the choice of generator may introduce lexical or structural regularities into the negative responses, which could partially influence the resulting steering alongside the ambiguity-related target signal. Finally, despite our manual re-labeling work, the ambiguous category in ClearVQA may still contain some degree of noise. In particular, we found that the original dataset contains a non-negligible number of examples that are not genuinely ambiguous with respect to the target, and some residual inconsistencies may therefore remain after filtering. 

Future work could extend this study along several directions. A natural next step would be to broaden the pool of evaluated models, both in number and in scale, to examine whether the observed trade-offs between steering, prompting, and \projname{} continue to hold in larger and more diverse VLMs. It would also be interesting to compare our inference-time steering approach against models explicitly fine-tuned for ambiguity-aware behavior in order to better understand how activation steering compares with task-specific training, as a complementary approach to ours.

\section*{Ethical Considerations}
Activation steering is a general-purpose technique for manipulating a model's internal representations, and like other similar techniques it its use therefore requires care.  The same mechanism used here could instead be applied to suppress safety-relevant behaviors, bypass alignment training, or push a model toward generating harmful, deceptive, or otherwise undesired content. Since steering works directly on activations instead of through the model's training objective, it can be more difficult to constrain than prompting-based interventions, which makes responsible use and careful validation of any steering direction particularly important before usage. We do not derive or apply any steering vectors seeking harmful behaviors in the present work, and we encourage future work to evaluate steering interventions for unintended side effects, before applying them in user-facing systems.

Separately, activation steering intervenes on the model's residual stream during generation, and such interventions are known to carry a risk of degrading output quality or increasing hallucination, independently of the specific target behavior. Future work applying steering more broadly should be aware that inducing a desired high-level behavior does not guarantee that the resulting generations remain factually grounded, and should evaluate reliability alongside behavior, rather than treating the latter as a sufficient indicator of a successful intervention.

We acknowledge the use of AI tools to support parts of the code development and writing refinement. All AI-generated outputs were carefully reviewed, and we confirm that we take full responsibility for the final content and results.

\section*{Acknowledgments}

We would like to thank several members of the MaiNLP lab, particularly Philipp Mondorf, Ryan Soh-Eun Shim, and Xinyuan Cheng, along with Arnisa Fazla, for their valuable feedback on earlier drafts of this paper. We also thank Leonardo Bertolazzi for his helpful ideas and stimulating discussions. Special thanks go to Jasmin Orth for her dedicated work on annotating ambiguity labels for the ClearVQA dataset, contributing to the validation of our LLM judge, and evaluating factual consistency. We acknowledge funding through the ERC Consolidator Grant DIALECT 101043235, and the German Research Foundation (DFG) project VariVLM (586887264).

\bibliography{custom}

@InProceedings{clark_molmo2_2026,
    author    = {Clark, Christopher and Zhang, Jieyu and Ma, Zixian and Park, Jae Sung and Tripathi, Rohun and Lee, Sangho and Salehi, Mohammadreza and Ren, Jason and Kim, Chris Dongjoo and Yang, Yinuo and Shao, Vincent and Yang, Yue and Huang, Weikai and Gao, Ziqi and Anderson, Taira and Zhang, Jianrui and Jain, Jitesh and Stoica, George and Farhadi, Ali and Krishna, Ranjay},
    title     = {Molmo2: Open Weights and Data for Vision-Language Models with Video Understanding and Grounding},
    booktitle = {Proceedings of the IEEE/CVF Conference on Computer Vision and Pattern Recognition (CVPR)},
    month     = {June},
    year      = {2026},
    pages     = {28652-28668}
}

@misc{wang_internvl35_2025,
	title = {{InternVL3}.5: {Advancing} {Open}-{Source} {Multimodal} {Models} in {Versatility}, {Reasoning}, and {Efficiency}},
	copyright = {Creative Commons Attribution 4.0 International},
	shorttitle = {{InternVL3}.5},
	url = {https://arxiv.org/abs/2508.18265},
    eprint = {2508.18265},
    archivePrefix = {arXiv},
    primaryClass = {cs.CV},
	doi = {10.48550/ARXIV.2508.18265},
	urldate = {2026-08-18},
	publisher = {arXiv},
	author = {Wang, Weiyun and Gao, Zhangwei and Gu, Lixin and Pu, Hengjun and Cui, Long and Wei, Xingguang and Liu, Zhaoyang and Jing, Linglin and Ye, Shenglong and Shao, Jie and Wang, Zhaokai and Chen, Zhe and Zhang, Hongjie and Yang, Ganlin and Wang, Haomin and Wei, Qi and Yin, Jinhui and Li, Wenhao and Cui, Erfei and Chen, Guanzhou and Ding, Zichen and Tian, Changyao and Wu, Zhenyu and Xie, Jingjing and Li, Zehao and Yang, Bowen and Duan, Yuchen and Wang, Xuehui and Hou, Zhi and Hao, Haoran and Zhang, Tianyi and Li, Songze and Zhao, Xiangyu and Duan, Haodong and Deng, Nianchen and Fu, Bin and He, Yinan and Wang, Yi and He, Conghui and Shi, Botian and He, Junjun and Xiong, Yingtong and Lv, Han and Wu, Lijun and Shao, Wenqi and Zhang, Kaipeng and Deng, Huipeng and Qi, Biqing and Ge, Jiaye and Guo, Qipeng and Zhang, Wenwei and Zhang, Songyang and Cao, Maosong and Lin, Junyao and Tang, Kexian and Gao, Jianfei and Huang, Haian and Gu, Yuzhe and Lyu, Chengqi and Tang, Huanze and Wang, Rui and Lv, Haijun and Ouyang, Wanli and Wang, Limin and Dou, Min and Zhu, Xizhou and Lu, Tong and Lin, Dahua and Dai, Jifeng and Su, Weijie and Zhou, Bowen and Chen, Kai and Qiao, Yu and Wang, Wenhai and Luo, Gen},
	year = {2025},
}

@misc{bai_qwen3-vl_2025,
	title = {Qwen3-{VL} {Technical} {Report}},
	copyright = {arXiv.org perpetual, non-exclusive license},
	url = {https://arxiv.org/abs/2511.21631},
    eprint = {2511.21631},
    archivePrefix = {arXiv},
    primaryClass = {cs.CV},
	doi = {10.48550/ARXIV.2511.21631},
	urldate = {2026-08-18},
	publisher = {arXiv},
	author = {Bai, Shuai and Cai, Yuxuan and Chen, Ruizhe and Chen, Keqin and Chen, Xionghui and Cheng, Zesen and Deng, Lianghao and Ding, Wei and Gao, Chang and Ge, Chunjiang and Ge, Wenbin and Guo, Zhifang and Huang, Qidong and Huang, Jie and Huang, Fei and Hui, Binyuan and Jiang, Shutong and Li, Zhaohai and Li, Mingsheng and Li, Mei and Li, Kaixin and Lin, Zicheng and Lin, Junyang and Liu, Xuejing and Liu, Jiawei and Liu, Chenglong and Liu, Yang and Liu, Dayiheng and Liu, Shixuan and Lu, Dunjie and Luo, Ruilin and Lv, Chenxu and Men, Rui and Meng, Lingchen and Ren, Xuancheng and Ren, Xingzhang and Song, Sibo and Sun, Yuchong and Tang, Jun and Tu, Jianhong and Wan, Jianqiang and Wang, Peng and Wang, Pengfei and Wang, Qiuyue and Wang, Yuxuan and Xie, Tianbao and Xu, Yiheng and Xu, Haiyang and Xu, Jin and Yang, Zhibo and Yang, Mingkun and Yang, Jianxin and Yang, An and Yu, Bowen and Zhang, Fei and Zhang, Hang and Zhang, Xi and Zheng, Bo and Zhong, Humen and Zhou, Jingren and Zhou, Fan and Zhou, Jing and Zhu, Yuanzhi and Zhu, Ke},
	year = {2025},
}

@misc{zou_representation_2023,
	title = {Representation {Engineering}: {A} {Top}-{Down} {Approach} to {AI} {Transparency}},
	copyright = {arXiv.org perpetual, non-exclusive license},
	shorttitle = {Representation {Engineering}},
	url = {https://arxiv.org/abs/2310.01405},
    eprint = {2310.01405},
    archivePrefix = {arXiv},
    primaryClass = {cs.LG},
	doi = {10.48550/ARXIV.2310.01405},
	urldate = {2026-08-18},
	publisher = {arXiv},
	author = {Zou, Andy and Phan, Long and Chen, Sarah and Campbell, James and Guo, Phillip and Ren, Richard and Pan, Alexander and Yin, Xuwang and Mazeika, Mantas and Dombrowski, Ann-Kathrin and Goel, Shashwat and Li, Nathaniel and Byun, Michael J. and Wang, Zifan and Mallen, Alex and Basart, Steven and Koyejo, Sanmi and Song, Dawn and Fredrikson, Matt and Kolter, J. Zico and Hendrycks, Dan},
	year = {2023},
}

@misc{yang_steering_2026,
	title = {Steering to {Say} {No}: {Configurable} {Refusal} via {Activation} {Steering} in {Vision} {Language} {Models}},
	copyright = {Creative Commons Attribution 4.0 International},
	shorttitle = {Steering to {Say} {No}},
	url = {https://arxiv.org/abs/2602.07013},
    eprint = {2602.07013},
    archivePrefix = {arXiv},
    primaryClass = {cs.AI},
	doi = {10.48550/ARXIV.2602.07013},
	urldate = {2026-08-13},
	publisher = {arXiv},
	author = {Yang, Jiaxi and Liu, Shicheng and Yang, Yuchen and Lee, Dongwon},
	year = {2026},
}

@inproceedings{sivakumar_steervlm_2025,
  title = {{SteerVLM}: Robust Model Control through Lightweight Activation Steering for Vision Language Models},
  author = {Sivakumar, Anushka and Zhang, Andrew and Abdul Hakim, Zaber Ibn and Thomas, Chris},
  booktitle = {Findings of the Association for Computational Linguistics: EMNLP 2025},
  year = {2025},
  pages = {23640--23665},
  publisher = {Association for Computational Linguistics},
  url = {https://aclanthology.org/2025.findings-emnlp.1285/},
  doi = {10.18653/v1/2025.findings-emnlp.1285}
}

@misc{arditi_refusal_2024,
	title = {Refusal in {Language} {Models} {Is} {Mediated} by a {Single} {Direction}},
	copyright = {arXiv.org perpetual, non-exclusive license},
	url = {https://arxiv.org/abs/2406.11717},
	doi = {10.48550/ARXIV.2406.11717},
    eprint = {2406.11717},
    archivePrefix = {arXiv},
    primaryClass = {cs.LG},
	urldate = {2026-08-13},
	publisher = {NeurIPS},
	author = {Arditi, Andy and Obeso, Oscar and Syed, Aaquib and Paleka, Daniel and Panickssery, Nina and Gurnee, Wes and Nanda, Neel},
	year = {2024},
}

@misc{park_linear_2023,
	title = {The {Linear} {Representation} {Hypothesis} and the {Geometry} of {Large} {Language} {Models}},
	copyright = {Creative Commons Attribution Non Commercial Share Alike 4.0 International},
	url = {https://arxiv.org/abs/2311.03658},
    eprint = {2311.03658},
    archivePrefix = {arXiv},
    primaryClass = {cs.LG},
	doi = {10.48550/ARXIV.2311.03658},
	urldate = {2026-08-13},
	publisher = {arXiv},
	author = {Park, Kiho and Choe, Yo Joong and Veitch, Victor},
	year = {2023},
}

@misc{alain_understanding_2016,
	title = {Understanding intermediate layers using linear classifier probes},
	copyright = {arXiv.org perpetual, non-exclusive license},
	url = {https://arxiv.org/abs/1610.01644},
	doi = {10.48550/ARXIV.1610.01644},
	urldate = {2026-08-11},
	publisher = {arXiv},
	author = {Alain, Guillaume and Bengio, Yoshua},
	year = {2016},
    eprint = {1610.01644},
    archivePrefix = {arXiv},
    primaryClass = {cs.LG},
}

@inproceedings{su_activation_2025,
	address = {Vienna, Austria},
	title = {Activation {Steering} {Decoding}: {Mitigating} {Hallucination} in {Large} {Vision}-{Language} {Models} through {Bidirectional} {Hidden} {State} {Intervention}},
	shorttitle = {Activation {Steering} {Decoding}},
	url = {https://aclanthology.org/2025.acl-long.634},
	doi = {10.18653/v1/2025.acl-long.634},
	language = {en},
	urldate = {2026-08-11},
	booktitle = {Proceedings of the 63rd {Annual} {Meeting} of the {Association} for {Computational} {Linguistics} ({Volume} 1: {Long} {Papers})},
	publisher = {Association for Computational Linguistics},
	author = {Su, Jingran and Chen, Jingfan and Li, Hongxin and Chen, Yuntao and Qing, Li and Zhang, Zhaoxiang},
	year = {2025},
	pages = {12964--12974},
}

@misc{yin_dynamic_2026,
	title = {Dynamic {Multimodal} {Activation} {Steering} for {Hallucination} {Mitigation} in {Large} {Vision}-{Language} {Models}},
	copyright = {arXiv.org perpetual, non-exclusive license},
	url = {https://arxiv.org/abs/2602.21704},
    eprint = {2602.21704},
    archivePrefix = {arXiv},
    primaryClass = {cs.CV},
	doi = {10.48550/ARXIV.2602.21704},
	urldate = {2026-08-11},
	publisher = {ICLR},
	author = {Yin, Jianghao and Chen, Qin and Chen, Kedi and Zhou, Jie and Wu, Xingjiao and He, Liang},
	year = {2026},
}

@inproceedings{rimsky_steering_2024,
	address = {Bangkok, Thailand},
	title = {Steering {Llama} 2 via {Contrastive} {Activation} {Addition}},
	url = {https://aclanthology.org/2024.acl-long.828},
	doi = {10.18653/v1/2024.acl-long.828},
	language = {en},
	urldate = {2026-08-11},
	booktitle = {Proceedings of the 62nd {Annual} {Meeting} of the {Association} for {Computational} {Linguistics} ({Volume} 1: {Long} {Papers})},
	publisher = {Association for Computational Linguistics},
	author = {Rimsky, Nina and Gabrieli, Nick and Schulz, Julian and Tong, Meg and Hubinger, Evan and Turner, Alexander},
	year = {2024},
	pages = {15504--15522},
}

@misc{turner_steering_2023,
	title = {Steering {Language} {Models} {With} {Activation} {Engineering}},
	copyright = {Creative Commons Attribution 4.0 International},
	url = {https://arxiv.org/abs/2308.10248},
    eprint = {2308.10248},
    archivePrefix = {arXiv},
    primaryClass = {cs.CL},
	doi = {10.48550/ARXIV.2308.10248},
	urldate = {2026-08-11},
	publisher = {arXiv},
	author = {Turner, Alexander Matt and Thiergart, Lisa and Leech, Gavin and Udell, David and Vazquez, Juan J. and Mini, Ulisse and MacDiarmid, Monte},
	year = {2023},
}

@inproceedings{min_ambigqa_2020,
	address = {Online},
	title = {{AmbigQA}: {Answering} {Ambiguous} {Open}-domain {Questions}},
	shorttitle = {{AmbigQA}},
	url = {https://aclanthology.org/2020.emnlp-main.466},
	doi = {10.18653/v1/2020.emnlp-main.466},
	language = {en},
	urldate = {2026-08-11},
	booktitle = {Proceedings of the 2020 {Conference} on {Empirical} {Methods} in {Natural} {Language} {Processing} ({EMNLP})},
	publisher = {Association for Computational Linguistics},
	author = {Min, Sewon and Michael, Julian and Hajishirzi, Hannaneh and Zettlemoyer, Luke},
	year = {2020},
	pages = {5783--5797},
}

@misc{jang_aqua_2026,
	title = {{AQuA}: {Toward} {Strategic} {Response} {Generation} for {Ambiguous} {Visual} {Questions}},
	copyright = {arXiv.org perpetual, non-exclusive license},
	shorttitle = {{AQuA}},
	url = {https://arxiv.org/abs/2603.07394},
	doi = {10.48550/ARXIV.2603.07394},
	language = {en},
	urldate = {2026-08-11},
	publisher = {arXiv},
    eprint = {2603.07394},
    archivePrefix = {arXiv},
    primaryClass = {cs.AI},
	author = {Jang, Jihyoung and Kim, Hyounghun},
	year = {2026},
}

@inproceedings{jian_teaching_2025,
	address = {Vienna, Austria},
	title = {Teaching {Vision}-{Language} {Models} to {Ask}: {Resolving} {Ambiguity} in {Visual} {Questions}},
	shorttitle = {Teaching {Vision}-{Language} {Models} to {Ask}},
	url = {https://aclanthology.org/2025.acl-long.182},
	doi = {10.18653/v1/2025.acl-long.182},
	language = {en},
	urldate = {2026-08-11},
	booktitle = {Proceedings of the 63rd {Annual} {Meeting} of the {Association} for {Computational} {Linguistics} ({Volume} 1: {Long} {Papers})},
	publisher = {Association for Computational Linguistics},
	author = {Jian, Pu and Yu, Donglei and Yang, Wen and Ren, Shuo and Zhang, Jiajun},
	year = {2025},
	pages = {3619--3638},
}

@misc{pres2024reliable,
      title={Towards Reliable Evaluation of Behavior Steering Interventions in LLMs}, 
      author={Itamar Pres and Laura Ruis and Ekdeep Singh Lubana and David Krueger},
      year={2024},
      eprint={2410.17245},
      archivePrefix={arXiv},
      primaryClass={cs.AI},
      url={https://arxiv.org/abs/2410.17245}, 
}

@misc{grattafiori2024llama3herdmodels,
      title={The Llama 3 Herd of Models}, 
      author={Aaron Grattafiori and Abhimanyu Dubey and Abhinav Jauhri and Abhinav Pandey and Abhishek Kadian and Ahmad Al-Dahle and Aiesha Letman and Akhil Mathur and Alan Schelten and Alex Vaughan and Amy Yang and Angela Fan and Anirudh Goyal and Anthony Hartshorn and Aobo Yang and Archi Mitra and Archie Sravankumar and Artem Korenev and Arthur Hinsvark and Arun Rao and Aston Zhang and Aurelien Rodriguez and Austen Gregerson and Ava Spataru and Baptiste Roziere and Bethany Biron and Binh Tang and Bobbie Chern and Charlotte Caucheteux and Chaya Nayak and Chloe Bi and Chris Marra and Chris McConnell and Christian Keller and Christophe Touret and Chunyang Wu and Corinne Wong and Cristian Canton Ferrer and Cyrus Nikolaidis and Damien Allonsius and Daniel Song and Danielle Pintz and Danny Livshits and Danny Wyatt and David Esiobu and Dhruv Choudhary and Dhruv Mahajan and Diego Garcia-Olano and Diego Perino and Dieuwke Hupkes and Egor Lakomkin and Ehab AlBadawy and Elina Lobanova and Emily Dinan and Eric Michael Smith and Filip Radenovic and Francisco Guzmán and Frank Zhang and Gabriel Synnaeve and Gabrielle Lee and Georgia Lewis Anderson and Govind Thattai and Graeme Nail and Gregoire Mialon and Guan Pang and Guillem Cucurell and Hailey Nguyen and Hannah Korevaar and Hu Xu and Hugo Touvron and Iliyan Zarov and Imanol Arrieta Ibarra and Isabel Kloumann and Ishan Misra and Ivan Evtimov and Jack Zhang and Jade Copet and Jaewon Lee and Jan Geffert and Jana Vranes and Jason Park and Jay Mahadeokar and Jeet Shah and Jelmer van der Linde and Jennifer Billock and Jenny Hong and Jenya Lee and Jeremy Fu and Jianfeng Chi and Jianyu Huang and Jiawen Liu and Jie Wang and Jiecao Yu and Joanna Bitton and Joe Spisak and Jongsoo Park and Joseph Rocca and Joshua Johnstun and Joshua Saxe and Junteng Jia and Kalyan Vasuden Alwala and Karthik Prasad and Kartikeya Upasani and Kate Plawiak and Ke Li and Kenneth Heafield and Kevin Stone and Khalid El-Arini and Krithika Iyer and Kshitiz Malik and Kuenley Chiu and Kunal Bhalla and Kushal Lakhotia and Lauren Rantala-Yeary and Laurens van der Maaten and Lawrence Chen and Liang Tan and Liz Jenkins and Louis Martin and Lovish Madaan and Lubo Malo and Lukas Blecher and Lukas Landzaat and Luke de Oliveira and Madeline Muzzi and Mahesh Pasupuleti and Mannat Singh and Manohar Paluri and Marcin Kardas and Maria Tsimpoukelli and Mathew Oldham and Mathieu Rita and Maya Pavlova and Melanie Kambadur and Mike Lewis and Min Si and Mitesh Kumar Singh and Mona Hassan and Naman Goyal and Narjes Torabi and Nikolay Bashlykov and Nikolay Bogoychev and Niladri Chatterji and Ning Zhang and Olivier Duchenne and Onur Çelebi and Patrick Alrassy and Pengchuan Zhang and Pengwei Li and Petar Vasic and Peter Weng and Prajjwal Bhargava and Pratik Dubal and Praveen Krishnan and Punit Singh Koura and Puxin Xu and Qing He and Qingxiao Dong and Ragavan Srinivasan and Raj Ganapathy and Ramon Calderer and Ricardo Silveira Cabral and Robert Stojnic and Roberta Raileanu and Rohan Maheswari and Rohit Girdhar and Rohit Patel and Romain Sauvestre and Ronnie Polidoro and Roshan Sumbaly and Ross Taylor and Ruan Silva and Rui Hou and Rui Wang and Saghar Hosseini and Sahana Chennabasappa and Sanjay Singh and Sean Bell and Seohyun Sonia Kim and Sergey Edunov and Shaoliang Nie and Sharan Narang and Sharath Raparthy and Sheng Shen and Shengye Wan and Shruti Bhosale and Shun Zhang and Simon Vandenhende and Soumya Batra and Spencer Whitman and Sten Sootla and Stephane Collot and Suchin Gururangan and Sydney Borodinsky and Tamar Herman and Tara Fowler and Tarek Sheasha and Thomas Georgiou and Thomas Scialom and Tobias Speckbacher and Todor Mihaylov and Tong Xiao and Ujjwal Karn and Vedanuj Goswami and Vibhor Gupta and Vignesh Ramanathan and Viktor Kerkez and Vincent Gonguet and Virginie Do and Vish Vogeti and Vítor Albiero and Vladan Petrovic and Weiwei Chu and Wenhan Xiong and Wenyin Fu and Whitney Meers and Xavier Martinet and Xiaodong Wang and Xiaofang Wang and Xiaoqing Ellen Tan and Xide Xia and Xinfeng Xie and Xuchao Jia and Xuewei Wang and Yaelle Goldschlag and Yashesh Gaur and Yasmine Babaei and Yi Wen and Yiwen Song and Yuchen Zhang and Yue Li and Yuning Mao and Zacharie Delpierre Coudert and Zheng Yan and Zhengxing Chen and Zoe Papakipos and Aaditya Singh and Aayushi Srivastava and Abha Jain and Adam Kelsey and Adam Shajnfeld and Adithya Gangidi and Adolfo Victoria and Ahuva Goldstand and Ajay Menon and Ajay Sharma and Alex Boesenberg and Alexei Baevski and Allie Feinstein and Amanda Kallet and Amit Sangani and Amos Teo and Anam Yunus and Andrei Lupu and Andres Alvarado and Andrew Caples and Andrew Gu and Andrew Ho and Andrew Poulton and Andrew Ryan and Ankit Ramchandani and Annie Dong and Annie Franco and Anuj Goyal and Aparajita Saraf and Arkabandhu Chowdhury and Ashley Gabriel and Ashwin Bharambe and Assaf Eisenman and Azadeh Yazdan and Beau James and Ben Maurer and Benjamin Leonhardi and Bernie Huang and Beth Loyd and Beto De Paola and Bhargavi Paranjape and Bing Liu and Bo Wu and Boyu Ni and Braden Hancock and Bram Wasti and Brandon Spence and Brani Stojkovic and Brian Gamido and Britt Montalvo and Carl Parker and Carly Burton and Catalina Mejia and Ce Liu and Changhan Wang and Changkyu Kim and Chao Zhou and Chester Hu and Ching-Hsiang Chu and Chris Cai and Chris Tindal and Christoph Feichtenhofer and Cynthia Gao and Damon Civin and Dana Beaty and Daniel Kreymer and Daniel Li and David Adkins and David Xu and Davide Testuggine and Delia David and Devi Parikh and Diana Liskovich and Didem Foss and Dingkang Wang and Duc Le and Dustin Holland and Edward Dowling and Eissa Jamil and Elaine Montgomery and Eleonora Presani and Emily Hahn and Emily Wood and Eric-Tuan Le and Erik Brinkman and Esteban Arcaute and Evan Dunbar and Evan Smothers and Fei Sun and Felix Kreuk and Feng Tian and Filippos Kokkinos and Firat Ozgenel and Francesco Caggioni and Frank Kanayet and Frank Seide and Gabriela Medina Florez and Gabriella Schwarz and Gada Badeer and Georgia Swee and Gil Halpern and Grant Herman and Grigory Sizov and Guangyi and Zhang and Guna Lakshminarayanan and Hakan Inan and Hamid Shojanazeri and Han Zou and Hannah Wang and Hanwen Zha and Haroun Habeeb and Harrison Rudolph and Helen Suk and Henry Aspegren and Hunter Goldman and Hongyuan Zhan and Ibrahim Damlaj and Igor Molybog and Igor Tufanov and Ilias Leontiadis and Irina-Elena Veliche and Itai Gat and Jake Weissman and James Geboski and James Kohli and Janice Lam and Japhet Asher and Jean-Baptiste Gaya and Jeff Marcus and Jeff Tang and Jennifer Chan and Jenny Zhen and Jeremy Reizenstein and Jeremy Teboul and Jessica Zhong and Jian Jin and Jingyi Yang and Joe Cummings and Jon Carvill and Jon Shepard and Jonathan McPhie and Jonathan Torres and Josh Ginsburg and Junjie Wang and Kai Wu and Kam Hou U and Karan Saxena and Kartikay Khandelwal and Katayoun Zand and Kathy Matosich and Kaushik Veeraraghavan and Kelly Michelena and Keqian Li and Kiran Jagadeesh and Kun Huang and Kunal Chawla and Kyle Huang and Lailin Chen and Lakshya Garg and Lavender A and Leandro Silva and Lee Bell and Lei Zhang and Liangpeng Guo and Licheng Yu and Liron Moshkovich and Luca Wehrstedt and Madian Khabsa and Manav Avalani and Manish Bhatt and Martynas Mankus and Matan Hasson and Matthew Lennie and Matthias Reso and Maxim Groshev and Maxim Naumov and Maya Lathi and Meghan Keneally and Miao Liu and Michael L. Seltzer and Michal Valko and Michelle Restrepo and Mihir Patel and Mik Vyatskov and Mikayel Samvelyan and Mike Clark and Mike Macey and Mike Wang and Miquel Jubert Hermoso and Mo Metanat and Mohammad Rastegari and Munish Bansal and Nandhini Santhanam and Natascha Parks and Natasha White and Navyata Bawa and Nayan Singhal and Nick Egebo and Nicolas Usunier and Nikhil Mehta and Nikolay Pavlovich Laptev and Ning Dong and Norman Cheng and Oleg Chernoguz and Olivia Hart and Omkar Salpekar and Ozlem Kalinli and Parkin Kent and Parth Parekh and Paul Saab and Pavan Balaji and Pedro Rittner and Philip Bontrager and Pierre Roux and Piotr Dollar and Polina Zvyagina and Prashant Ratanchandani and Pritish Yuvraj and Qian Liang and Rachad Alao and Rachel Rodriguez and Rafi Ayub and Raghotham Murthy and Raghu Nayani and Rahul Mitra and Rangaprabhu Parthasarathy and Raymond Li and Rebekkah Hogan and Robin Battey and Rocky Wang and Russ Howes and Ruty Rinott and Sachin Mehta and Sachin Siby and Sai Jayesh Bondu and Samyak Datta and Sara Chugh and Sara Hunt and Sargun Dhillon and Sasha Sidorov and Satadru Pan and Saurabh Mahajan and Saurabh Verma and Seiji Yamamoto and Sharadh Ramaswamy and Shaun Lindsay and Shaun Lindsay and Sheng Feng and Shenghao Lin and Shengxin Cindy Zha and Shishir Patil and Shiva Shankar and Shuqiang Zhang and Shuqiang Zhang and Sinong Wang and Sneha Agarwal and Soji Sajuyigbe and Soumith Chintala and Stephanie Max and Stephen Chen and Steve Kehoe and Steve Satterfield and Sudarshan Govindaprasad and Sumit Gupta and Summer Deng and Sungmin Cho and Sunny Virk and Suraj Subramanian and Sy Choudhury and Sydney Goldman and Tal Remez and Tamar Glaser and Tamara Best and Thilo Koehler and Thomas Robinson and Tianhe Li and Tianjun Zhang and Tim Matthews and Timothy Chou and Tzook Shaked and Varun Vontimitta and Victoria Ajayi and Victoria Montanez and Vijai Mohan and Vinay Satish Kumar and Vishal Mangla and Vlad Ionescu and Vlad Poenaru and Vlad Tiberiu Mihailescu and Vladimir Ivanov and Wei Li and Wenchen Wang and Wenwen Jiang and Wes Bouaziz and Will Constable and Xiaocheng Tang and Xiaojian Wu and Xiaolan Wang and Xilun Wu and Xinbo Gao and Yaniv Kleinman and Yanjun Chen and Ye Hu and Ye Jia and Ye Qi and Yenda Li and Yilin Zhang and Ying Zhang and Yossi Adi and Youngjin Nam and Yu and Wang and Yu Zhao and Yuchen Hao and Yundi Qian and Yunlu Li and Yuzi He and Zach Rait and Zachary DeVito and Zef Rosnbrick and Zhaoduo Wen and Zhenyu Yang and Zhiwei Zhao and Zhiyu Ma},
      year={2024},
      eprint={2407.21783},
      archivePrefix={arXiv},
      primaryClass={cs.AI},
      url={https://arxiv.org/abs/2407.21783}, 
}

@misc{lee2025programmingrefusalconditionalactivation,
      title={Programming Refusal with Conditional Activation Steering}, 
      author={Bruce W. Lee and Inkit Padhi and Karthikeyan Natesan Ramamurthy and Erik Miehling and Pierre Dognin and Manish Nagireddy and Amit Dhurandhar},
      year={2025},
      eprint={2409.05907},
      archivePrefix={ICLR},
      primaryClass={cs.LG},
      url={https://arxiv.org/abs/2409.05907}, 
}

@inproceedings{testoni_racquet_2025,
  title={{RACQUET}: Unveiling the Dangers of Overlooked Referential Ambiguity in Visual {LLM}s},
  author={Testoni, Alberto and Plank, Barbara and Fern{\'a}ndez, Raquel},
  booktitle={Proceedings of the 2025 Conference on Empirical Methods in Natural Language Processing (EMNLP)},
  publisher = {Association for Computational Linguistics},
  pages={23627--23647},
  year={2025}
}

@inproceedings{stengel_eskin_chicken_2023,
  title={Why Did the Chicken Cross the Road? Rephrasing and Analyzing Ambiguous Questions in {VQA}},
  author={Stengel-Eskin, Elias and Guallar-Blasco, Jimena and Zhou, Yi and Van Durme, Benjamin},
  booktitle={Proceedings of the 61st Annual Meeting of the Association for Computational Linguistics (Volume 1: Long Papers)},
  pages={10220--10237},
  year={2023},
  address = {Toronto, Canada},
  publisher = {Association for Computational Linguistics},
}

@misc{li2024inferencetimeinterventionelicitingtruthful,
      title={Inference-Time Intervention: Eliciting Truthful Answers from a Language Model}, 
      author={Kenneth Li and Oam Patel and Fernanda Viégas and Hanspeter Pfister and Martin Wattenberg},
      year={2024},
      eprint={2306.03341},
      archivePrefix={arXiv},
      primaryClass={cs.LG},
      url={https://arxiv.org/abs/2306.03341}, 
}

@misc{yang2025qwen3technicalreport,
      title={Qwen3 Technical Report}, 
      author={An Yang and Anfeng Li and Baosong Yang and Beichen Zhang and Binyuan Hui and Bo Zheng and Bowen Yu and Chang Gao and Chengen Huang and Chenxu Lv and Chujie Zheng and Dayiheng Liu and Fan Zhou and Fei Huang and Feng Hu and Hao Ge and Haoran Wei and Huan Lin and Jialong Tang and Jian Yang and Jianhong Tu and Jianwei Zhang and Jianxin Yang and Jiaxi Yang and Jing Zhou and Jingren Zhou and Junyang Lin and Kai Dang and Keqin Bao and Kexin Yang and Le Yu and Lianghao Deng and Mei Li and Mingfeng Xue and Mingze Li and Pei Zhang and Peng Wang and Qin Zhu and Rui Men and Ruize Gao and Shixuan Liu and Shuang Luo and Tianhao Li and Tianyi Tang and Wenbiao Yin and Xingzhang Ren and Xinyu Wang and Xinyu Zhang and Xuancheng Ren and Yang Fan and Yang Su and Yichang Zhang and Yinger Zhang and Yu Wan and Yuqiong Liu and Zekun Wang and Zeyu Cui and Zhenru Zhang and Zhipeng Zhou and Zihan Qiu},
      year={2025},
      eprint={2505.09388},
      archivePrefix={arXiv},
      primaryClass={cs.CL},
      url={https://arxiv.org/abs/2505.09388}, 
}

@misc{prasad_rephrase_2023,
  title = {Rephrase, Augment, Reason: Visual Grounding of Questions for Vision-Language Models},
  author = {Prasad, Archiki and Stengel-Eskin, Elias and Bansal, Mohit},
  year = {2023},
  eprint = {2310.05861},
  archivePrefix = {arXiv},
  primaryClass = {cs.CV},
}

@InProceedings{Nam_2025_ICCV,
    author    = {Nam, Heejeong and Ahn, Jinwoo and Ka, Keummin and Chung, Jiwan and Yu, Youngjae},
    title     = {{VAGUE}: Visual Contexts Clarify Ambiguous Expressions},
    booktitle = {Proceedings of the IEEE/CVF International Conference on Computer Vision (ICCV)},
    month     = {October},
    year      = {2025},
    pages     = {1537-1547}
}

@incollection{grice1975,
  author    = {H. P. Grice},
  title     = {Logic and Conversation},
  booktitle = {Speech Acts: Syntax and Semantics Volume 3},
  editor    = {Peter Cole and Jerry L. Morgan},
  pages     = {41--58},
  publisher = {Academic Press},
  address   = {New York},
  year      = {1975},
  doi       = {10.1163/9789004368811_003}
}

@inproceedings{
marks2024the,
title={The Geometry of Truth: Emergent Linear Structure in Large Language Model Representations of True/False Datasets},
author={Samuel Marks and Max Tegmark},
booktitle={First Conference on Language Modeling},
year={2024},
url={https://openreview.net/forum?id=aajyHYjjsk}
}

@InProceedings{pmlr-clip-radford21a,
  title = 	 {Learning Transferable Visual Models From Natural Language Supervision},
  author =       {Radford, Alec and Kim, Jong Wook and Hallacy, Chris and Ramesh, Aditya and Goh, Gabriel and Agarwal, Sandhini and Sastry, Girish and Askell, Amanda and Mishkin, Pamela and Clark, Jack and Krueger, Gretchen and Sutskever, Ilya},
  booktitle = 	 {Proceedings of the 38th International Conference on Machine Learning},
  pages = 	 {8748--8763},
  year = 	 {2021},
  editor = 	 {Meila, Marina and Zhang, Tong},
  volume = 	 {139},
  series = 	 {Proceedings of Machine Learning Research},
  month = 	 {18--24 Jul},
  publisher =    {PMLR},
  url = 	 {https://proceedings.mlr.press/v139/radford21a.html}
}

@InProceedings{Zhai_2023_siglip,
    author    = {Zhai, Xiaohua and Mustafa, Basil and Kolesnikov, Alexander and Beyer, Lucas},
    title     = {Sigmoid Loss for Language Image Pre-Training},
    booktitle = {Proceedings of the IEEE/CVF International Conference on Computer Vision (ICCV)},
    month     = {October},
    year      = {2023},
    pages     = {11975-11986}
}

@InProceedings{pmlr-blip-li22n,
  title = 	 {{BLIP}: Bootstrapping Language-Image Pre-training for Unified Vision-Language Understanding and Generation},
  author =       {Li, Junnan and Li, Dongxu and Xiong, Caiming and Hoi, Steven},
  booktitle = 	 {Proceedings of the 39th International Conference on Machine Learning},
  pages = 	 {12888--12900},
  year = 	 {2022},
  editor = 	 {Chaudhuri, Kamalika and Jegelka, Stefanie and Song, Le and Szepesvari, Csaba and Niu, Gang and Sabato, Sivan},
  volume = 	 {162},
  series = 	 {Proceedings of Machine Learning Research},
  month = 	 {17--23 Jul},
  publisher =    {PMLR},
  url = 	 {https://proceedings.mlr.press/v162/li22n.html}
}

@InProceedings{pmlr-vilt-kim21k,
  title = 	 {{ViLT}: Vision-and-Language Transformer Without Convolution or Region Supervision},
  author =       {Kim, Wonjae and Son, Bokyung and Kim, Ildoo},
  booktitle = 	 {Proceedings of the 38th International Conference on Machine Learning},
  pages = 	 {5583--5594},
  year = 	 {2021},
  editor = 	 {Meila, Marina and Zhang, Tong},
  volume = 	 {139},
  series = 	 {Proceedings of Machine Learning Research},
  month = 	 {18--24 Jul},
  publisher =    {PMLR},
  url = 	 {https://proceedings.mlr.press/v139/kim21k.html}
}

@InProceedings{Antol_2015_ICCV,
author = {Antol, Stanislaw and Agrawal, Aishwarya and Lu, Jiasen and Mitchell, Margaret and Batra, Dhruv and Zitnick, C. Lawrence and Parikh, Devi},
title = {{VQA}: Visual Question Answering},
booktitle = {Proceedings of the IEEE International Conference on Computer Vision (ICCV)},
month = {December},
year = {2015}
}

@inproceedings{shim-etal-2026-linear,
    title = "Linear Script Representations in Speech Foundation Models Enable Zero-Shot Transliteration",
    author = "Shim, Ryan Soh-Eun  and
      Choi, Kwanghee  and
      Chang, Kalvin  and
      Hsu, Ming-Hao  and
      Eichin, Florian  and
      Wu, Zhizheng  and
      Suhr, Alane  and
      Hedderich, Michael A.  and
      Harwath, David  and
      Mortensen, David R.  and
      Plank, Barbara",
    editor = "Liakata, Maria  and
      Moreira, Viviane P.  and
      Zhang, Jiajun  and
      Jurgens, David",
    booktitle = "Findings of the {A}ssociation for {C}omputational {L}inguistics: {ACL} 2026",
    month = jul,
    year = "2026",
    address = "San Diego, California, United States",
    publisher = "Association for Computational Linguistics",
    url = "https://aclanthology.org/2026.findings-acl.464/",
    doi = "10.18653/v1/2026.findings-acl.464",
    pages = "9532--9544",
    ISBN = "979-8-89176-395-1"
}

@inproceedings{
dao2024flashattention,
title={FlashAttention-2: Faster Attention with Better Parallelism and Work Partitioning},
author={Tri Dao},
booktitle={The Twelfth International Conference on Learning Representations},
year={2024},
url={https://openreview.net/forum?id=mZn2Xyh9Ec}
}

\appendix

\section{Dataset Adaptation}
\label{app:dataset_adaptation}
\begin{figure}[th!]
    \centering
    \includegraphics[width=\linewidth]{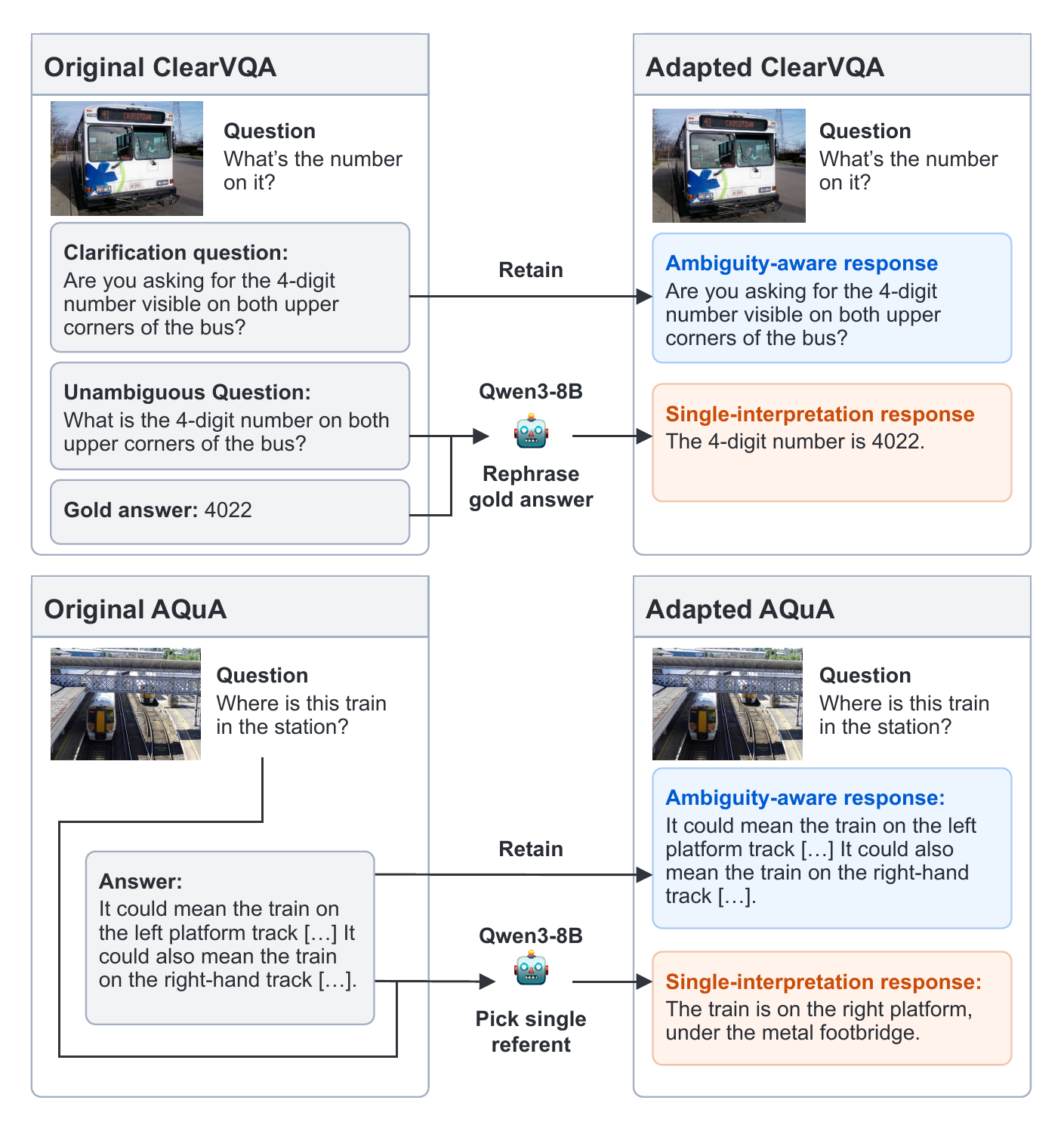}
    \caption{Adaptation of ClearVQA and AQuA into ambiguity-aware and single-interpretation responses.
}
    \label{fig:dataset_adaptation}
\end{figure}

We use ClearVQA \citep{jian_teaching_2025} and AQuA \citep{jang_aqua_2026}, two English-language VQA datasets targeting referential ambiguity in general-domain visual scenes, as the basis for our experiments. Because these two benchmarks are designed for open-ended generation or fine-tuning rather than representation intervention, we adapt instances from ClearVQA and AQuA to construct contrastive pairs. Each adapted instance consists of a shared image and question paired with two alternative responses: a positive, ambiguity-aware response (requesting clarification or enumerating plausible referents) and a negative, single-interpretation response (arbitrarily selecting a single referent). Both adaptations were run using Qwen3-8B \citep{yang2025qwen3technicalreport} to rephrase the original datasets. Figure~\ref{fig:dataset_adaptation} illustrates this adaptation process for both benchmarks, with details for each benchmark provided below.

To ensure the quality of the generated single-interpretation responses, two authors of the paper qualitatively inspected a random sample of 50 responses per dataset (100 in total). We verified that each response commits to a single referent rather than preserving the ambiguity, while remaining faithful to the original content. Out of the 100 inspected responses, only one failed to commit to a single interpretation, yielding a 99\% compliance rate.

\begin{figure}[t]
\centering
\begin{promptbox}{ClearVQA adaptation prompt}
\textbf{System}

You are a helpful assistant. Your task is to rephrase a short VQA answer into a single, complete, natural sentence. Use the question as context so the sentence is self-contained. Output ONLY the rephrased sentence, no explanation beyond the sentence itself.

\medskip
\textbf{User}

Question: \texttt{\{original question\}}

Short answer: \texttt{\{short answer\}}

Rephrased answer:
\end{promptbox}
\caption{Prompt used to construct ClearVQA single-interpretation responses.}
\label{fig:clearvqa-construction-prompt}
\end{figure}

\paragraph{ClearVQA.}
The dataset was constructed from originally unambiguous VQA examples, each consisting of an image, a question, and a short ground-truth answer. The authors then use an LLM-based pipeline to introduce ambiguity by removing or blurring information from the question and to generate a corresponding clarification question. Each ClearVQA instance therefore contains the original unambiguous question and gold answer alongside an ambiguous question and clarification question. We augmented clarification questions with an ambiguity-aware response and unambiguous questions with a single-interpretation response. For our adaptation, we use the clarification question as the ambiguity-aware response and prompt Qwen3-8B \citep{yang2025qwen3technicalreport} to rewrite the gold answer as a complete sentence conditioned on the original unambiguous question, yielding the corresponding single-interpretation response (see Figure \ref{fig:clearvqa-construction-prompt}).

\paragraph{AQuA.}
The dataset organizes image-question pairs into four levels of increasing ambiguity (from 0 to 3).
We use Level~2, where multiple plausible referents remain and the gold responses enumerate them, making it a natural setting for our enumeration task.
For each Level~2 instance, the gold response serves as the ambiguity-aware response. For the corresponding single-referent response, we prompt Qwen3-8B  \citep{yang2025qwen3technicalreport} with the question and gold response and ask it to answer using only one of the referents listed in the gold response, as shown in Figure \ref{fig:aqua-construction-prompt}.

\begin{figure}[t]
\centering
\begin{promptbox}{AQuA adaptation prompt}
\textbf{System}

You are a helpful assistant. Your task is to generate a plausible sounding answer that mentions ONLY ONE of the elements or interpretations in the gold answer, completely ignoring the others. Reply with ONLY the bad answer text, phrased as one complete sentence, no explanation beyond the sentence itself.

\medskip
\textbf{User}

Question: \texttt{\{question\}}

Golden answer: \texttt{\{gold answer\}}

Bad answer:
\end{promptbox}
\caption{Prompt used to construct AQuA single-interpretation responses.}
\label{fig:aqua-construction-prompt}
\end{figure}

\begin{figure*}[th!]
    \centering
    \includegraphics[width=\textwidth]{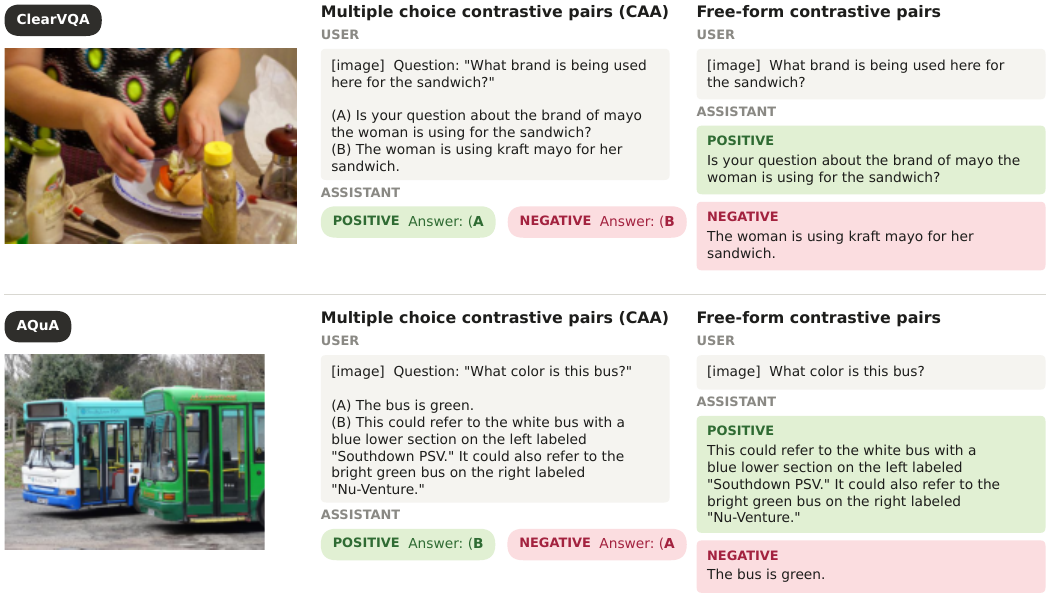}
    \caption{Examples of multiple-choice (CAA) and free-form contrastive pairs for ClearVQA and AQuA.}
    \label{fig:examples_contrastive_pairs}
\end{figure*}

\section{Contrastive Pair Examples}
\label{app:constrastive_constructions}

Figure \ref{fig:examples_contrastive_pairs} shows representative contrastive pairs for ClearVQA and AQuA under the multiple-choice (CAA) and free-form (FF) constructions. For CAA, positive and negative sequences differ only in the final answer-letter token. The ambiguity-aware response is assigned to option A or B in balanced proportions to avoid systematic answer-position bias.
For FF, the corresponding responses are appended directly to the shared image-question context, resulting in sequences that may differ in length and wording.

\section{Ambiguity Labels Derivation}
\label{app:human_ambiguity_annotation}

We derive binary ambiguity labels separately for ClearVQA and AQuA.

\subsection{ClearVQA Ambiguity Labels}

For ClearVQA, although the dataset is intended to contain ambiguous questions, manual inspection revealed that a substantial number of instances in the released referential subset are not genuinely ambiguous with respect to the corresponding image. We therefore manually label each example as \textit{ambiguous}, \textit{non-ambiguous}, \textit{uncertain}, or \textit{poor quality}, retaining only the first two categories. 

Following \citet{jian_teaching_2025}, we consider a question referentially ambiguous when its referring expression does not uniquely identify the intended referent. For each example, one annotator was shown the question and corresponding image, together with the original unambiguous question and its gold answer from the ClearVQA dataset as reference. Examples were presented individually in random order.

For each example, the annotator assigned one of four labels: \textit{ambiguous} if the intended referent could not be uniquely identified from the question and image; \textit{non-ambiguous} if the intended referent remained identifiable and the question could be answered directly; \textit{uncertain} if the annotator could not confidently distinguish between these two cases; and \textit{poor quality} if the example did not support a reliable judgment, e.g. because the question did not fit the image or was nonsensical.

\begin{figure}[t!]
    \centering
    \includegraphics[width=\linewidth]{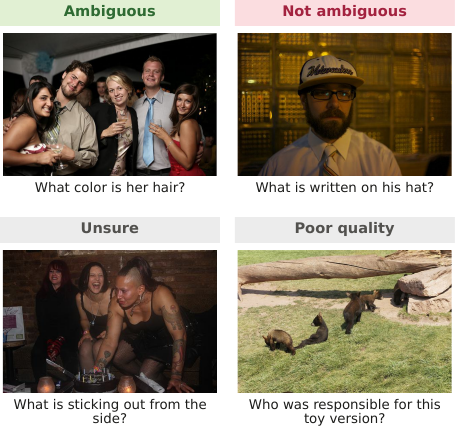}
    \caption{Examples for the human ambiguity labels.}
    \label{fig:ambiguity_labels}
\end{figure}

Of the 1,928 unique questions, 995 (51.6\%) were labeled \textit{ambiguous}, 633 (32.8\%) \textit{non-ambiguous}, 114 (5.9\%) \textit{uncertain}, and 186 (9.6\%) \textit{poor quality}. We retain the 1,628 examples labeled ambiguous or non-ambiguous and discard the remaining categories. Figure~\ref{fig:ambiguity_labels} shows one representative example for each of the four labels.

The annotation was carried out by a master's student in Computational Linguistics, employed as a student assistant and compensated according to national standards. The annotator was informed of the intended research use of the annotations and agreed to perform the task.

\subsection{AQuA Ambiguity Labels}
The AQuA dataset \citep{jang_aqua_2026} organizes examples into four levels of increasing ambiguity. Level 0 contains unambiguous questions; at Level 1, the referent is underspecified in the question but can still be recovered from the image; at Level 2, multiple plausible referents remain and the gold response enumerates them; and Level 3 requires clarification because the intended referent cannot be resolved. We map Levels 0 and 1 to non-ambiguous and Level 2 to ambiguous.
We do not use Level 3 because it targets clarification behavior, which we cover with ClearVQA. In addition, ClearVQA retains the original unambiguous question and gold answer from which each ambiguous question was derived, allowing us to construct a single-interpretation response for the negative contrastive pair. AQuA does not provide such an unambiguous counterpart for Level 3.

\begin{figure}[h!]
\centering
\begin{promptbox}{Clarification Prompt}
\textit{[image]}

\textit{$\langle$question$\rangle$}

\medskip

\textbf{Instruction:}
If the question is ambiguous, ask a clarifying question instead of answering.
\end{promptbox}
\caption{Generation instruction used for ClearVQA.}
\label{fig:clarification_instruction}
\end{figure}

\begin{figure}[h!]
\centering
\begin{promptbox}{Enumeration Prompt}
\textit{[image]}

\textit{$\langle$question$\rangle$}

\medskip

\textbf{Instruction:}
If the question is referentially ambiguous, explicitly list each possible
referent instead of assuming a single interpretation.
\end{promptbox}
\caption{Generation instruction used for AQuA.}
\label{fig:enumeration_instruction}
\end{figure}

\begin{figure}[h!]
\centering
\begin{promptbox}{Clarification judge template}

\smallskip

You are an evaluator. You are given the user's original question
and a short text generated by a model in response to it. Your task is to decide whether that text contains a clarification question for the user or not. Only mark as yes if the question is truly a clarification question, i.e. it asks the user to clarify, specify, or disambiguate what they meant, and not a random question that is not asking for clarification. In addition, only mark as yes if a generated clarification truly refers to and is related to the user's original question. A question that instead offers to perform a different or additional task, or that asks whether the user would like some other or further information, is not a clarification question. Mark as no if the text already commits to a referent before asking for clarification. Answer with exactly one word: ``yes'' if the text contains such a clarification question, or ``no'' if it does not. Do not explain your answer.

\medskip

\begin{center}
\textit{[Four few-shot demonstrations are inserted here]}
\end{center}

\medskip

\smallskip

\textbf{User's original question:}

\texttt{\{question\}}

\smallskip

\textbf{Text to evaluate:}

\texttt{\{response\}}

\smallskip

\end{promptbox}
\caption{Prompt template used by the clarification judge.}
\label{fig:clarification_judge}
\end{figure}

\section{Generation and Evaluation Prompts}
\label{app:prompt_templates}

\subsection{Generation Instructions}
\label{app:generation_instructions}

All generation settings use the image and question as input. In
\emph{Baseline} and \emph{Steer}, no additional instruction is
provided. In \emph{Prompt}, \emph{Prompt+Steer} and \projname{}, we additionally provide the dataset-specific instruction for \emph{clarification} and \emph{enumeration} shown in Figure \ref{fig:clarification_instruction} and Figure \ref{fig:enumeration_instruction}, respectively. These generation prompts are the same for CAA and FF; the two constructions differ only in how the steering vectors are obtained.

\subsection{LLM-Judge Prompts}
\label{app:llm_judge_prompts}

We use Meta-Llama-3.1-8B-Instruct~\citep{grattafiori2024llama3herdmodels} as a few-shot judge for
assessing the presence of clarification and enumeration behavior in model generations. For each task, the conversation
consists of the task description, four demonstrations as alternating user and assistant turns, and the item to be evaluated as the final user turn. Figures~\ref{fig:clarification_judge} and
\ref{fig:enumeration_judge} show the two judge prompts for the clarification and enumeration judges, while
Figures~\ref{fig:clarification_fewshot} and
\ref{fig:enumeration_fewshot} show their corresponding few-shot
demonstrations.

\begin{figure}[th!]
\centering
\begin{promptbox}{Enumeration judge template}

\textbf{System prompt}

\smallskip

You are an evaluator. You are given a question a user asked about an image and a short text generated by a model in response to it.
Your task is to decide whether the response enumerates multiple possible referents of the referring expression in the question.

Mark yes only if both of the following hold: (1) the response presents two or more distinct objects, people, or regions as possible referents of the expression in the question, and (2) it gives a separate answer to the question for each of those referents. The response does not need to explicitly state that the question is ambiguous.

Mark no if the response commits to a single referent, even if it gives a detailed answer, and even if it describes other objects or people as part of the surroundings or as supporting detail.

Mark no if the response lists, counts, or describes several objects or people without giving separate answers to the question for them as possible referents.

Answer with exactly one word: yes or no. Do not explain your answer.

\medskip
\textbf{Few-shot demonstrations}

\smallskip

\begin{center}
\textit{[Four few-shot demonstrations are inserted here]}
\end{center}

\medskip
\textbf{Final user turn}

\smallskip

QUESTION: \texttt{\{question\}}

\smallskip

MODEL GENERATION: \texttt{\{response\}}

\end{promptbox}
\caption{Prompt template used by the enumeration judge.}
\label{fig:enumeration_judge}
\end{figure}

\begin{judgepromptbox}{Clarification judge: few-shot demonstrations}

\textbf{Demonstration 1}

\smallskip

\textit{User's original question:}  
What color is his jacket?

\smallskip

\textit{Text to evaluate:}  
Based on the image provided, there are four men standing in front of a bench. The man on the far right is wearing a jacket.

The color of his jacket is \textit{green}. It appears to be a dark green or forest green color, and he is also wearing a beanie.

\smallskip

\textit{Assistant:}
\texttt{no}

\medskip
\textbf{Demonstration 2}

\smallskip

\textit{User's original question:}  
What color is his jacket?

\smallskip

\textit{Text to evaluate:}  
The question is ambiguous because it refers to ``his jacket'' without specifying which person is being referred to. There are four men in the image, each wearing a different jacket:

\begin{itemize}
    \item The man on the far left is wearing a blue jacket.
    \item The man second from the left is wearing a gray jacket.
    \item The man third from the left is wearing a dark blue or black jacket.
    \item The man on the far right is wearing a green jacket.
\end{itemize}

Could you please specify which person you are referring to?

\smallskip

\textit{Assistant:}
\texttt{yes}

\medskip
\textbf{Demonstration 3}

\smallskip

\textit{User's original question:}  
How many times does it appear here?

\smallskip

\textit{Text to evaluate:}  
The Apple logo appears \textit{5 times} in the image.

Here's a breakdown of where each one is located:

\begin{enumerate}
    \item On the white keyboard box (top right corner, partially visible).
    \item On the white keyboard (center, on the spacebar).
    \item On the white mouse (center right, on the top surface).
    \item On the left user guide (top center, on the cover).
    \item On the right user guide (top center, on the cover).
\end{enumerate}

The logo is also visible on the gray box in the background, but it's partially obscured and not clearly distinct from the other items. The most clearly visible and distinct instances are the 5 listed above.

\smallskip

\textit{Assistant:}
\texttt{no}

\medskip
\textbf{Demonstration 4}

\smallskip

\textit{User's original question:}  
How many times does it appear here?

\smallskip

\textit{Text to evaluate:}  
The question ``How many times does it appear here?'' is ambiguous without specifying what ``it'' refers to.

Could you clarify:
\begin{itemize}
    \item What specific object, text, or feature are you asking about?
    \item Are you referring to the Apple logo, the word ``Wireless,'' the keyboard, the mouse, or something else?
\end{itemize}

Once clarified, I can give you an accurate count.

\smallskip

\textit{Assistant:}
\texttt{yes}

\end{judgepromptbox}

\begingroup
\captionof{figure}{Few-shot demonstrations provided to the clarification judge.}
\label{fig:clarification_fewshot}
\endgroup

\begin{judgepromptbox}{Enumeration judge: few-shot demonstrations}

\textbf{Demonstration 1}

\smallskip

\textit{QUESTION:}  
What is sitting on top of it?

\smallskip

\textit{MODEL GENERATION:}  
The question is referentially ambiguous because ``it'' could refer to either:

\begin{itemize}
    \item The wooden bench (which has a cat sitting on it)
    \item The wooden piano (which also has items on its top surface)
\end{itemize}

Therefore, the answer is:

\begin{itemize}
    \item A cat is sitting on top of the bench.
    \item The piano also has sheet music and bowls on its top surface.
\end{itemize}

\smallskip

\textit{Assistant:} \texttt{yes}

\medskip
\textbf{Demonstration 2}

\smallskip

\textit{QUESTION:}  
What is sitting on top of it?

\smallskip

\textit{MODEL GENERATION:}  
Based on the image provided, a cat is sitting on top of the piano bench.

Here are the details:

\begin{itemize}
    \item The cat is a long-haired, gray tabby with a striped pattern.
    \item It is lying down comfortably on the wooden bench, which is positioned in front of the upright piano.
\end{itemize}

So, the animal sitting on top of the bench is a cat.

\smallskip

\textit{Assistant:} \texttt{no}

\medskip
\textbf{Demonstration 3}

\smallskip

\textit{QUESTION:}  
What color is that house behind him?

\smallskip

\textit{MODEL GENERATION:}  
The image contains multiple houses in the background, so the color is not uniquely determined by a single referent. Here are the possible referents:

\begin{itemize}
    \item A large, multi-story house on the left side of the image, which has sections of gray siding and white trim around windows and balconies.
    \item A house to the right of the first one, which is white.
\end{itemize}

\smallskip

\textit{Assistant:} \texttt{yes}

\medskip
\textbf{Demonstration 4}

\smallskip

\textit{QUESTION:}  
What color is that house behind him?

\smallskip

\textit{MODEL GENERATION:}  
Based on the image provided, the house directly behind the man is a gray color. It appears to be a multi-story residential building with a gray exterior and a dark roof that has skylights. There is also a white house visible to the right of the gray one.

\smallskip

\textit{Assistant:} \texttt{no}

\end{judgepromptbox}

\begingroup
\captionof{figure}{Few-shot demonstrations provided to the enumeration judge.}
\label{fig:enumeration_fewshot}
\endgroup

\begin{table*}[t]
\centering
\small

\begin{tabular*}{0.95\textwidth}{@{\extracolsep{\fill}}llcccccc@{}}
\toprule
& &
\multicolumn{2}{c}{\textbf{Steer}} &
\multicolumn{2}{c}{\textbf{Prompt+Steer}$^{+}$} &
\multicolumn{2}{c}{\textbf{Prompt+Steer}$^{-}$} \\
\cmidrule(lr){3-4}
\cmidrule(lr){5-6}
\cmidrule(lr){7-8}

\textbf{Dataset} & \textbf{Model}
& $k=1$ & $k=5$
& $k=1$ & $k=5$
& $k=1$ & $k=5$ \\
\midrule

\multicolumn{8}{@{}l}{\textbf{CAA}} \\
\addlinespace[2pt]

\multirow{3}{*}{ClearVQA}
& InternVL3.5 & 1.5 & 0.6 & 2.5 & 0.6 & $-4.0$ & $-0.7$ \\
& Molmo2      & 1.5 & 0.6 & 1.0 & 0.2 & $-2.5$ & $-0.6$ \\
& Qwen3-VL    & 0.5 & 0.2 & 0.5 & 0.3 & $-2.0$ & $-0.6$ \\

\cmidrule(lr){2-8}

\multirow{3}{*}{AQuA}
& InternVL3.5 & 2.5 & 0.5 & 2.5 & 0.7 & $-3.0$ & $-0.5$ \\
& Molmo2      & 0.5 & 0.7 & 1.5 & 0.6 & $-3.5$ & $-0.4$ \\
& Qwen3-VL    & 1.0 & 0.4 & 1.5 & 0.5 & $-0.5^{\dagger}$ & $-0.1$ \\

\midrule

\multicolumn{8}{@{}l}{\textbf{Free-form (FF)}} \\
\addlinespace[2pt]

\multirow{3}{*}{ClearVQA}
& InternVL3.5 & 0.75 & 0.20 & 1.00 & 0.25 & $-1.75$ & $-0.35$ \\
& Molmo2      & 0.50 & 0.15 & 0.75 & 0.20 & $-1.25$ & $-0.35$ \\
& Qwen3-VL    & 0.25 & 0.10 & 0.25 & 0.10 & $-1.00$ & $-0.40$ \\

\cmidrule(lr){2-8}

\multirow{3}{*}{AQuA}
& InternVL3.5 & 1.25 & 0.35 & 1.00 & 0.20 & $-1.00$ & $-0.25$ \\
& Molmo2      & 1.00 & 0.30 & 0.75 & 0.15 & $-0.75$ & $-0.15$ \\
& Qwen3-VL    & 0.50 & 0.10 & 0.25 & 0.10 & $-0.50$ & $-0.10$ \\

\bottomrule
\end{tabular*}

\caption{Selected steering coefficients for each condition.
Prompt+Steer$^{+}$ and Prompt+Steer$^{-}$ denote positive and negative steering with the task-specific prompt, respectively.
$\dagger$ For Qwen3-VL on AQuA with CAA and $k=1$, all negative coefficients exceeded the 5\% degeneration threshold; we therefore use $\alpha=-0.5$ (6.3\% degeneration rate).}
\label{tab:selected_coefficients}
\end{table*}

\section{Layer and Coefficient Selection}
\label{app:layer_coeff_selection}

\paragraph{Layer selection.}
We consider $k \in \{1,5\}$, where $k$ denotes the number of consecutive decoder layers being steered. Following prior work, we sweep all decoder layers on the development sets using coefficients $+1$ and $-1$ 
\citep{rimsky_steering_2024}. However, instead of selecting layers from multiple-choice answer probabilities, we evaluate their effect directly on open-ended generation, motivated by recent work showing that steering effects measured in multiple-choice settings may not transfer reliably to open-ended behavior \citep{pres2024reliable}. For each layer, we compare the target-behavior rate under positive and negative steering, where the target behavior is clarification on ClearVQA and enumeration on AQuA. We measure this rate using the LLM-based judge described in Section~\ref{sec:metrics}. To keep the layer choice fixed across comparisons and limit per-condition tuning, we pool these results for layer selection. We average the resulting difference across models and datasets and, for each $k$, we select the contiguous $k$-layer window with the largest average difference in target-behavior rate. This gives layer~21 for $k=1$ and layers~20-24 for $k=5$. Selecting layers separately on ClearVQA and AQuA yields the same choices as the pooled selection. We use these same layers for vectors extracted from both free-form and multiple-choice contrastive pairs.

\begin{table}[t]
\centering
\small
\setlength{\tabcolsep}{4pt}
\begin{tabular}{@{}ccl@{}}
\toprule
\textbf{Vectors} & $k$ & \textbf{Coefficient magnitudes} \\
\midrule
CAA & 1 & 0.5, 1, 1.5, 2, 2.5, 3, 3.5, 4 \\
CAA & 5 & 0.1, 0.2, 0.3, 0.4, 0.5, 0.6, 0.7, 0.8 \\
FF  & 1 & 0.25, 0.5, 0.75, 1, 1.25, 1.5, 1.75, 2 \\
FF  & 5 & 0.05, 0.1, 0.15, 0.2, 0.25, 0.3, 0.35, 0.4 \\
\bottomrule
\end{tabular}
\caption{Coefficient ranges used for the development-set sweeps. The same sweep ranges are used for both clarification and enumeration behaviors.}
\label{tab:coef_grid}
\end{table}

\paragraph{Coefficient selection.}
After selecting the intervention layers, we tune the steering coefficient on each dataset's development set separately for each model, contrastive construction, number of steered layers, and generation setting.
To make steering magnitudes comparable across the two target behaviors, we norm-match the vectors on a per-layer basis, following~\citet{rimsky_steering_2024}, using ClearVQA as the reference and rescaling the corresponding vectors derived from AQuA accordingly.
To make coefficient sweeps comparable across different numbers of steered layers, we scale the coefficient applied at each layer by $1/k$ where $k$ is the number of steered layers.
Because vectors extracted from the free-form construction have larger norms, we scale the FF coefficient magnitudes by 0.5 relative to CAA. Table~\ref{tab:coef_grid} reports the coefficient grids used for each construction and $k$.
For our \textit{Steer} generation setup, we tune only positive coefficients, since this condition tests whether steering can \textit{elicit} the target behavior. For \textit{Prompt+Steer}, we tune both signs, which are subsequently also reused by \projname{} to promote ambiguity-aware behavior on inputs predicted as ambiguous and suppress it on inputs predicted as non-ambiguous.
For positive steering, we select the coefficient with the highest target-behavior rate. For negative steering, we select the coefficient with the lowest rate.
In both cases we consider only coefficients with at most 5\% degenerate generations (as defined in Section~\ref{sec:metrics}).
If all coefficients exceed the 5\% degeneration threshold, we select the coefficient with the lowest degeneration rate. Table~\ref{tab:selected_coefficients} reports the final selected coefficients for all experiments.

\section{Additional Ambiguity Classifier Analyses}
\label{app:alt_classifiers}

Our main experiments use a linear probe over Qwen3-VL residual-stream representations to predict input ambiguity and gate steering for all three evaluated VLMs. We further analyze this classifier setup by examining (i) how ambiguity decodability varies across Qwen3-VL layers, (ii) whether performance depends on which of the three evaluated VLMs provides the gating representation, and (iii) how the Qwen3-VL probe compares with classifiers based on models outside the set of steered VLMs.

\subsection{Layer-wise Ambiguity Decodability}
\begin{figure}[t]
    \centering
    \includegraphics[width=\linewidth]{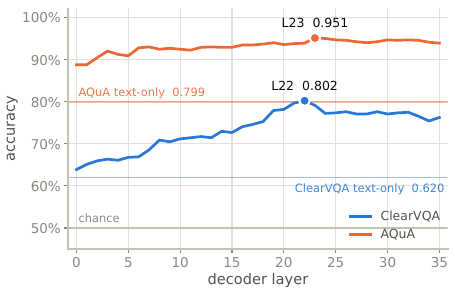}
    \caption{Layer-wise ambiguity decodability from Qwen3-VL residual-stream representations. Curves show 5-fold cross-validation accuracy on the train split.}
    \label{fig:classifier_layer}
\end{figure}

To examine how ambiguity information is represented across Qwen3-VL layers, we fit a separate logistic-regression probe to the final prompt-token residual representation at each decoder layer. For each layer, we measure 5-fold cross-validation accuracy on the training split and select the layer with the highest cross-validation accuracy for held-out evaluation. Figure \ref{fig:classifier_layer} shows the resulting layer-wise performance. On ClearVQA, accuracy increases from 63.9\% at layer 0 to 80.2\% at the selected layer 22. In contrast, AQuA ambiguity is already highly decodable at layer 0 (88.8\%), with accuracy reaching 95.1\% at the selected layer 23. At these selected layers, held-out test accuracy is 77.2\% on ClearVQA and 93.8\% on AQuA.

The different layer-wise patterns suggest that ClearVQA benefits more from deeper representations for ambiguity classification, whereas much of the ambiguity signal in AQuA is already accessible in early representations suggesting a possible presence of stronger surface-level linguistic cues in AQuA.

\subsection{Choice of gating VLM}

We next assess whether classification performance depends on the VLM providing the gating representation. We fit logistic-regression probes to Qwen3-VL, Molmo2, and InternVL3.5 residual-stream representations, selecting the layer for each model by 5-fold cross-validation on the training split.

\begin{table}[t]
\centering
\small
\begin{tabular}{lcccc}
\toprule
& \multicolumn{2}{c}{ClearVQA} & \multicolumn{2}{c}{AQuA} \\
\cmidrule(lr){2-3} \cmidrule(lr){4-5}
Gating VLM & CV & Test & CV & Test \\
\midrule
Qwen3-VL    & \textbf{80.2} & 77.2          & \textbf{95.1} & \textbf{93.8} \\
Molmo2      & 79.1          & \textbf{78.3} & 94.8          & \textbf{93.8} \\
InternVL3.5 & 77.2          & 78.0          & 94.0          & 93.2 \\
\bottomrule
\end{tabular}
\caption{Ambiguity-classification accuracy (\%) for the three VLMs used in our steering experiments.}
\label{tab:gating_vlm}
\end{table}

As shown in Table \ref{tab:gating_vlm}, Qwen3-VL achieves the highest cross-validation accuracy on both datasets (80.2\% on ClearVQA; 95.1\% on AQuA), motivating its use as the shared gating VLM. Held-out accuracy remains similar across all three models, ranging from 77.2-78.3\% on ClearVQA and 93.2-93.8\% on AQuA, suggesting that performance is not particularly sensitive to the choice of gating VLM.

\begin{figure}[t]
    \centering
    \includegraphics[width=\linewidth]{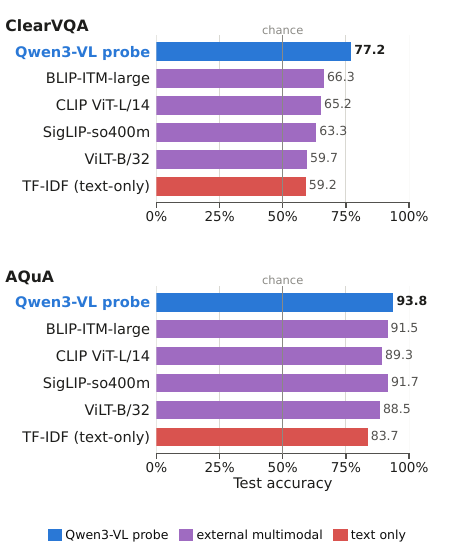}
    \caption{Held-out ambiguity-classification accuracy of the Qwen3-VL probe and external classifiers on ClearVQA and AQuA.}
    \label{fig:external_classifiers}
\end{figure}

\subsection{External classifier comparison}

We finally compare the Qwen3-VL probe with classifiers based on models outside the three VLMs considered for steering. These include a text-only TF-IDF classifier and multimodal classifiers based on CLIP \citep{pmlr-clip-radford21a}, SigLIP \citep{Zhai_2023_siglip}, BLIP \citep{pmlr-blip-li22n}, and ViLT \citep{pmlr-vilt-kim21k} representations. For each alternative classifier, model selection is performed using 5-fold cross-validation on the training split, followed by evaluation on the held-out test set.

As shown in Figure \ref{fig:external_classifiers}, the Qwen3-VL probe achieves the highest accuracy on both datasets, with 77.2\% on ClearVQA and 93.8\% on AQuA. External multimodal classifiers reach 59.7-66.3\% on ClearVQA and 88.5-91.7\% on AQuA, while the text-only classifier reaches 59.2\% and 83.7\%, respectively.

\section{Additional Human Evaluation Details}
\label{app:human_validation}

\subsection{Details on LLM-Judge Validation}
\label{app:llm_judge_validation}

We validate the clarification and enumeration judgments on 100 generated responses from each dataset. The same responses were independently labeled by two annotators: one internally hired master’s student in computational linguistics and one author of the paper. Annotators were shown the input question and generated response and assigned a binary label using the same criteria as the LLM judge. The LLM-judge labels were not shown during annotation.

We selected the validation samples to ensure that both the presence and absence of the target behavior were sufficiently represented. For ClearVQA, we uniformly sampled 100 ambiguous development examples from Qwen3-VL-8B generations under the \textit{Prompt} generation setting, a setting in which clarification occurs with non-trivial frequency.
For AQuA, we stratified the sample by ambiguity level, using 50 ambiguous Level-2 examples and 50 non-ambiguous examples, split evenly between Levels 0 and 1. Half of the non-ambiguous examples were generated with CAA steering ($\alpha=0.25$) so that the sample also included cases in which enumeration was elicited unnecessarily. All validation examples were disjoint from the few-shot demonstrations used by the LLM judge.

Table \ref{tab:judge_validation} reports the full pairwise agreement scores. We distinguish the paper author from the internally hired annotator to make the comparison with the LLM judge explicit. LLM–author agreement is one percentage point higher than LLM–student agreement on both datasets, indicating only a minor difference across annotators.

\begin{table}[t]
\centering
\small
\begin{tabular}{llcc}
\toprule
Dataset & Comparison & Agreement & $\kappa$ \\
\midrule
ClearVQA
& Author--Student & 97\% & 0.931 \\
& LLM--Author     & 91\% & 0.804 \\
& LLM--Student    & 90\% & 0.779 \\
\midrule
AQuA
& Author--Student & 91\% & 0.795 \\
& LLM--Author     & 90\% & 0.778 \\
& LLM--Student    & 89\% & 0.760 \\
\bottomrule
\end{tabular}
\caption{Pairwise agreement for the human validation of the LLM-Judge. Agreement is raw label agreement; $\kappa$ is Cohen's kappa.}
\label{tab:judge_validation}
\end{table}

\subsection{Details on Factual Consistency Evaluation}
\label{app:factual_consistenc_annotation}

We evaluate factual consistency on 100 AQuA test items for each of the three models, for a total of 300 \projname{} responses. For each model, the sample contains 50 ambiguous and 50 non-ambiguous inputs. The same set of 100 items is used across models. We evaluate the responses produced by \projname{} under positive or negative steering based on the ambiguity classifier.

The samples are annotated by an internally hired master’s student in computational linguistics. For each response, the annotator is shown the image, question, and generated response; the gold answer is not shown. The annotation instructions are adapted from the factual-consistency criterion and evaluation protocol of AQuA \citep{jang_aqua_2026} and are shown in Figure \ref{fig:factual_instruction}. Responses are presented in shuffled order.

Table \ref{tab:factual-consistency} reports the resulting factual-consistency rates. Overall, 77.3\% of AmbiSteer responses are judged factually consistent, with rates of 71.3\% on ambiguous and 83.3\% on non-ambiguous inputs.

\begin{figure}[t]
\centering
\begin{promptbox}{Factual-Consistency Annotation \\ Instructions}
Evaluate whether the generated response is factually consistent with the image.

Ignore style, grammar, and completeness.

Mark \textbf{PASS} if the factual claims in the response are supported by what is visible in the image, even if the response does not mention everything or omits relevant details.

Mark \textbf{FAIL} if the response mentions something not present in the image, contradicts the image, or introduces unsupported details.
\end{promptbox}
\caption{Factual-consistency instructions shown to the human annotator.}
\label{fig:factual_instruction}
\end{figure}

\begin{table}[t]
\centering
\small
\setlength{\tabcolsep}{4pt}
\renewcommand{\arraystretch}{1.08}
\begin{tabular}{lccc}
\toprule
\textbf{Model} & \textbf{Amb.} & \textbf{Non-amb.} & \textbf{Overall} \\
\midrule
InternVL3.5 & 80.0 & 92.0 & 86.0 \\
Molmo2      & 66.0 & 70.0 & 68.0 \\
Qwen3-VL    & 68.0 & 88.0 & 78.0 \\
\midrule
Average     & 71.3 & 83.3 & 77.3 \\
\bottomrule
\end{tabular}
\caption{Factual consistency (\%) of \projname{} responses on AQuA.}
\label{tab:factual-consistency}
\end{table}

\section{Additional Steering Results}
\label{app:free_form}

The main text focuses on CAA vectors applied across five layers ($k=5$, L20–24). Here, we report results for the three remaining configurations: CAA with single-layer steering ($k=1$, L21), free-form (FF) steering with $k=5$, and FF steering with $k=1$. For easier comparison, we also include the CAA $k=5$ results
reported in the main text. The layers and all coefficients are selected on the development sets following the procedure in Appendix \ref{app:layer_coeff_selection}.

\subsection{CAA with Single-Layer Steering}

Figure \ref{fig:app_grid_elicitation} (second row) reports the results for CAA when steering only layer 21. As in the five-layer setting, steering alone has limited effects in most cases, whereas combining steering with prompting increases the target-behavior rate over prompting alone for every model and dataset. The gains range from 16.3 to 54.7 percentage points, showing that the main pattern does not depend on intervening across multiple layers.

\projname{} with $k=1$ also reproduces the behavioral-accuracy pattern observed for CAA with $k=5$ in the main text. As shown in Table~\ref{tab:app-accuracy}, it outperforms Prompt+Steer for all six model-dataset settings, increasing mean behavioral accuracy from 60.8\% to 78.2\%. It also outperforms the shared Prompt baseline in every setting. For comparison, the table includes the CAA $k=5$ results reported in the main text.

\subsection{Free-form Steering}

\begin{figure*}[th!]
    \centering
    \includegraphics[width=\textwidth]{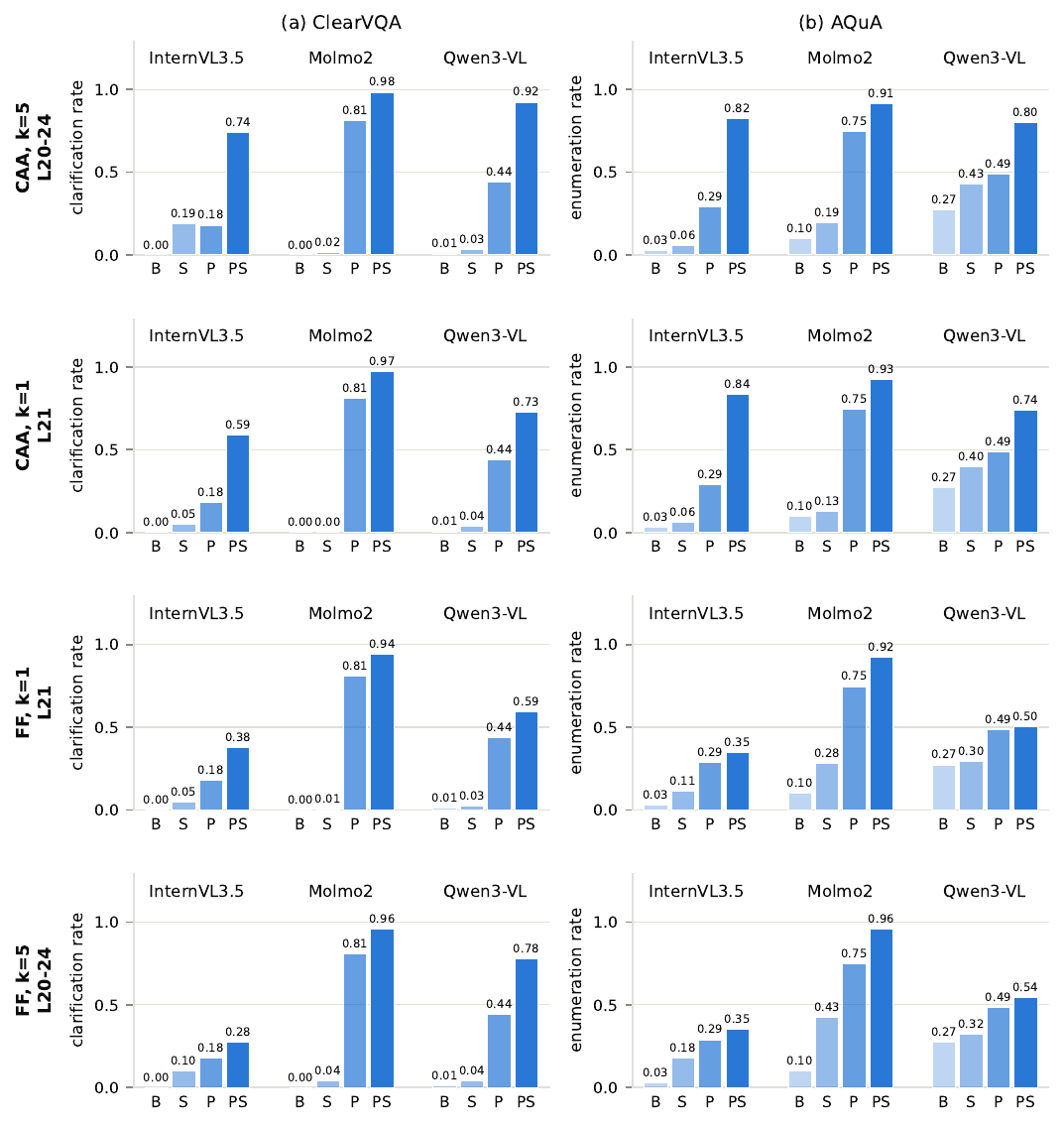}
    \caption{ Target-behavior rates on ambiguous test inputs across steering configurations. Rows show CAA with $k=5$, CAA with $k=1$, FF with $k=1$, and FF with $k=5$; columns show ClearVQA and AQuA. The CAA $k=5$ row repeats the main-text results for comparison. Single-layer steering uses L21, while five-layer steering uses L20--24 B = Baseline, S = Steer, P = Prompt, and PS = Prompt+Steer.
}
    \label{fig:app_grid_elicitation}
\end{figure*}

We next evaluate steering vectors obtained with the free-form (FF) construction. Figure~\ref{fig:app_grid_elicitation} (bottom two rows) reports results for FF with $k=1$ and $k=5$, respectively. For both settings, Prompt+Steer increases the target-behavior rate over Prompt alone for every model and dataset. The magnitude of this increase is more model-dependent than for CAA: FF is particularly effective for Molmo2, while CAA yields larger increases for InternVL3.5 and Qwen3-VL in most settings. This model-dependent pattern is similar for $k=1$ and $k=5$.

The benefit of \projname{} extends to FF. 
It improves behavioral accuracy over Prompt+Steer for every model and dataset under both $k=1$ and $k=5$ (Table~\ref{tab:app-accuracy}), and also exceeds the shared Prompt baseline in every setting. Mean behavioral accuracy increases from 58.9\% to 70.9\% for $k=1$ and from 58.4\% to 71.6\% for $k=5$. Thus, the improvement from conditioning is not specific to the CAA construction, although its magnitude depends on the model and steering configuration.

\begin{table*}[t]
\centering
\footnotesize
\setlength{\tabcolsep}{3.2pt}
\renewcommand{\arraystretch}{1.08}

\begin{tabular}{llc@{\hspace{5pt}}cc@{\hspace{5pt}}cc@{\hspace{5pt}}cc@{\hspace{5pt}}cc}
\toprule
& &
\multirow{2}{*}{\textbf{Prompt}} &
\multicolumn{2}{c}{\textbf{CAA $k{=}5$}} &
\multicolumn{2}{c}{\textbf{CAA $k{=}1$}} &
\multicolumn{2}{c}{\textbf{FF $k{=}1$}} &
\multicolumn{2}{c}{\textbf{FF $k{=}5$}} \\
\cmidrule(lr){4-5}
\cmidrule(lr){6-7}
\cmidrule(lr){8-9}
\cmidrule(lr){10-11}
\textbf{Dataset} & \textbf{Model} &
& PS & AS & PS & AS & PS & AS & PS & AS \\
\midrule

\multirow{3}{*}{ClearVQA}
& InternVL3.5
& 55.2
& 60.7 & \ws{30}{\textbf{73.2}}
& 57.7 & \ws{20}{\underline{67.3}}
& 58.7 & \ws{10}{61.2}
& 55.0 & \ws{10}{59.0} \\

& Molmo2
& 60.7
& 55.5 & \ws{40}{75.2}
& 54.7 & \ws{50}{\underline{76.2}}
& 53.0 & \ws{50}{75.5}
& 53.0 & \ws{50}{\textbf{76.5}} \\

& Qwen3-VL
& 63.0
& 51.3 & \ws{50}{\textbf{73.8}}
& 59.7 & \ws{30}{72.0}
& 59.3 & \ws{30}{70.2}
& 60.0 & \ws{30}{\underline{73.7}} \\

\midrule

\multirow{3}{*}{AQuA}
& InternVL3.5
& 60.7
& 67.3 & \ws{40}{\underline{86.5}}
& 66.5 & \ws{50}{\textbf{87.2}}
& 61.0 & \ws{10}{65.5}
& 62.2 & \ws{10}{64.5} \\

& Molmo2
& 66.0
& 55.5 & \ws{40}{74.7}
& 59.8 & \ws{60}{\textbf{90.7}}
& 56.5 & \ws{60}{86.3}
& 53.5 & \ws{60}{\underline{87.7}} \\

& Qwen3-VL
& 64.2
& 63.5 & \ws{30}{\textbf{78.5}}
& 66.7 & \ws{20}{\underline{76.0}}
& 65.2 & \ws{10}{66.5}
& 66.7 & \ws{10}{68.3} \\

\bottomrule
\end{tabular}

\caption{
Behavioral accuracy (\%) on the balanced test sets for Prompt,
Prompt+Steer (PS), and \projname{} (AS) across steering
configurations. Prompt is shared across configurations because no activation steering is applied. Shading shows the gain of AS over the corresponding PS setting; darker cells indicate larger gains.
\textbf{Bold} and \underline{underlined} values indicate the best and second-best result in each row, respectively.}
\label{tab:app-accuracy}
\end{table*}

\section{Qualitative Examples}
\label{app:qualitative_examples}

\begin{figure*}[th!]
    \centering
    \includegraphics[width=\textwidth]{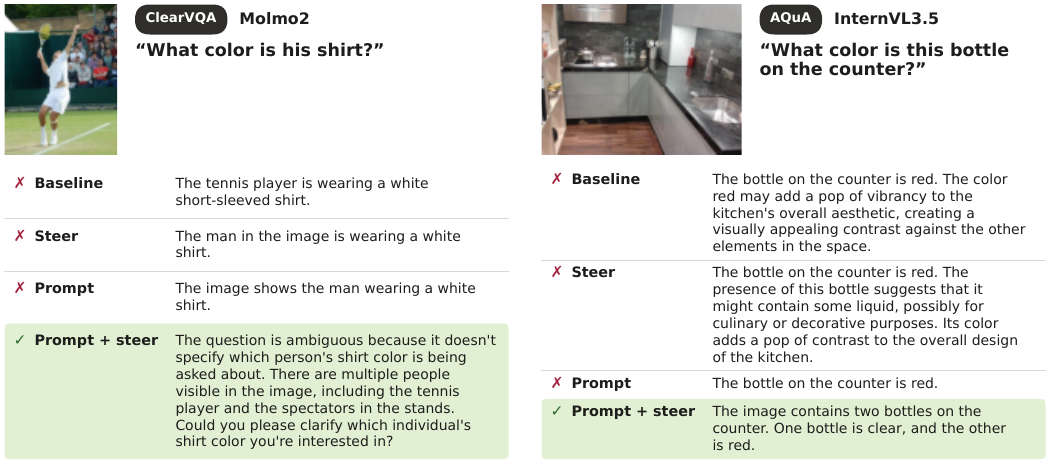}
    \caption{Example generations under different generation settings on ClearVQA and AQuA.}
    \label{fig:examples_4_settings}
\end{figure*}

\begin{figure*}[th!]
    \centering
    \includegraphics[width=\textwidth]{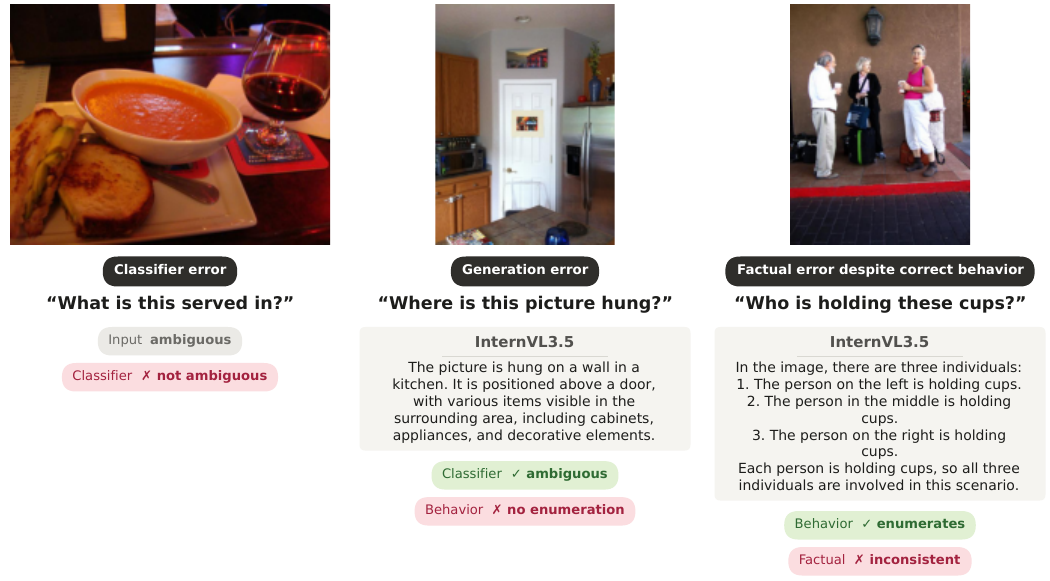}
    \caption{Examples of the failure modes analyzed in our error analysis: ambiguity classification errors, generation errors with incorrect behavior despite correct ambiguity classification, and factual errors despite correct enumeration behavior.}
    \label{fig:examples_errors}
\end{figure*}

Figure~\ref{fig:examples_4_settings} shows example generations under our different generation settings for ClearVQA and AQuA. Figure~\ref{fig:examples_errors} shows qualitative examples of the different error types analyzed in Section ~\ref{sec:error_analysis}: ambiguity classification errors, generation errors (despite correct ambiguity classification), and factual errors (despite correct enumeration behavior).

\section{Implementation Details}
\label{app:implementation_details}

We use the Hugging Face implementations\footnote{\url{https://huggingface.co/models}} of all models with their default chat templates and preprocessing, and load all models in bfloat16 with FlashAttention-2 \citep{dao2024flashattention}. All generations use greedy decoding with a maximum of 512 new tokens. Steering-specific hyperparameters are set as described in Section~\ref{sec:hyperparameters} and Appendix~\ref{app:layer_coeff_selection}. All experiments are run on NVIDIA A100 80GB GPUs.

\end{document}